%% file: main.tex
\documentclass[conference]{IEEEtran}
\usepackage{times}

\usepackage[numbers]{natbib}
\usepackage{multicol}
\usepackage[bookmarks=true, colorlinks=true, allcolors=blue]{hyperref}
\usepackage{graphicx}
\usepackage{algorithm}
\usepackage[labelfont=bf,format=plain]{caption}
\usepackage{algpseudocode}
\usepackage{courier}
\usepackage{amsmath}
\usepackage{comment}
\usepackage{amsfonts}
\usepackage{makecell}
\usepackage{booktabs}
\usepackage{multirow}
\usepackage{colortbl}
\usepackage[many]{tcolorbox}
\usepackage{multicol} 
\usepackage{float}
\usepackage{placeins}
\let\labelindent\relax
\usepackage{enumitem}

\newcommand\blfootnote[1]{%
  \begin{NoHyper}%
  \begingroup
  \renewcommand\thefootnote{}\footnote{#1}%
  \addtocounter{footnote}{-1}%
  \endgroup
  \end{NoHyper}%
}

\begin{document}

\title{CLIP-RT: Learning Language-Conditioned Robotic Policies from Natural Language Supervision}

\author{Gi-Cheon Kang$^{1*}$ \; Junghyun Kim$^{1*}$ \; Kyuhwan Shim$^1$ \; Jun Ki Lee$^{1\dagger}$ \;  Byoung-Tak Zhang$^{1,2\dagger}$ \vspace*{0.1cm}\\
$^1$Seoul National University \;\; $^2$Tommoro Robotics\\ \textbf{\url{https://clip-rt.github.io}}}



%

\maketitle

\begin{abstract}
Teaching robots desired skills in real-world environments remains challenging, especially for non-experts. A key bottleneck is that collecting robotic data often requires expertise or specialized hardware, limiting accessibility and scalability. We posit that natural language offers an intuitive and accessible interface for robot learning. To this end, we study two aspects: (1) enabling non-experts to collect robotic data through natural language supervision (\textit{e.g.,} ``move the arm to the right'') and (2) training robot policies directly from this supervision. Specifically, we introduce a data collection framework that collects robot demonstrations based on natural language supervision and further augments these demonstrations. We then present CLIP-RT, a new vision-language-action (VLA) model that learns language-conditioned visuomotor policies from this supervision. CLIP-RT adapts the pretrained CLIP model and learns to predict language-based motion primitives via contrastive imitation learning. We train CLIP-RT on the Open X-Embodiment dataset and finetune it on in-domain data collected by our framework. In real-world evaluations, CLIP-RT demonstrates strong capabilities in learning novel manipulation skills, outperforming OpenVLA (7B parameters) by 24\% in average success rates, while using 7x fewer parameters (1B). We further assess CLIP-RT's capabilities in few-shot generalization and collaborative scenarios involving large pretrained models or humans. In simulated environments, CLIP-RT also yields strong performance, achieving a 93.1\% average success rate on the LIBERO benchmark with an inference throughput of 163 Hz.
\end{abstract}

\IEEEpeerreviewmaketitle
\input{sections/01_intro}
\input{sections/02_related_works}

\input{sections/03_approach}

\input{sections/04_experiment}

\input{sections/05_libero}
\input{sections/06_limitation}

\input{sections/07_ack}
\bibliographystyle{plainnat}
\bibliography{references}
\clearpage
\input{sections/08_appendix}
\end{document}

%% file: sections/01_intro.tex
\section{Introduction}
Building robots that can understand natural language instructions and perform various real-world tasks is a long-standing goal of robotics and artificial intelligence. The research community has studied such robots in various domains, such as robotic manipulation~\citep{mees2022calvin,brohan2022rt,brohan2023rt}, navigation~\citep{anderson2018vision,das2018embodied,thomason2020vision,ku2020room}, and other instructions-following tasks~\citep{shridhar2020alfred,padmakumar2022teach}. \blfootnote{* equal contribution; $\dagger$ equal advising.} 

One key challenge for intelligent robots is grounding natural language to vision and action, bridging the abstraction gap between natural language instruction and visuomotor control in real-world tasks. Prior works on robotic manipulation have addressed this challenge by training language-conditioned policies, primarily through imitation learning~\citep{stepputtis2020language,lynch2020language,shridhar2022cliport,jang2022bc,mees2022calvin,brohan2022rt,brohan2023rt}. This line of research has shown remarkable success as large amounts of robotic data become available~\cite{padalkar2023open}. However, even state-of-the-art models~\cite{brohan2023rt,padalkar2023open,belkhale2024rt,kim24openvla} trained in large-scale robot data struggle to easily expand their set of manipulation skills for a wide range of real-world tasks. We argue that a major bottleneck lies in how robot demonstrations are typically collected. Specifically, obtaining real-world robot demonstration data often requires expertise in robot control or access to specialized hardware, such as teleoperation or virtual reality (VR) systems~\citep{xiao2022robotic,fu2024mobile}. This barrier severely limits accessibility, restricting the number of participants and environments from which data can be gathered. Consequently, this limited accessibility inherently hinders both the scalability (the volume of data) and the diversity (the range of scenarios and behaviors recorded) of the resulting datasets. We thus ask: \emph{how can non-experts train robotic policies without relying on specialized expertise or devices for data collection?}
\begin{figure}[t!]
\centering
\label{fig:teaser}
\includegraphics[width=0.45\textwidth]{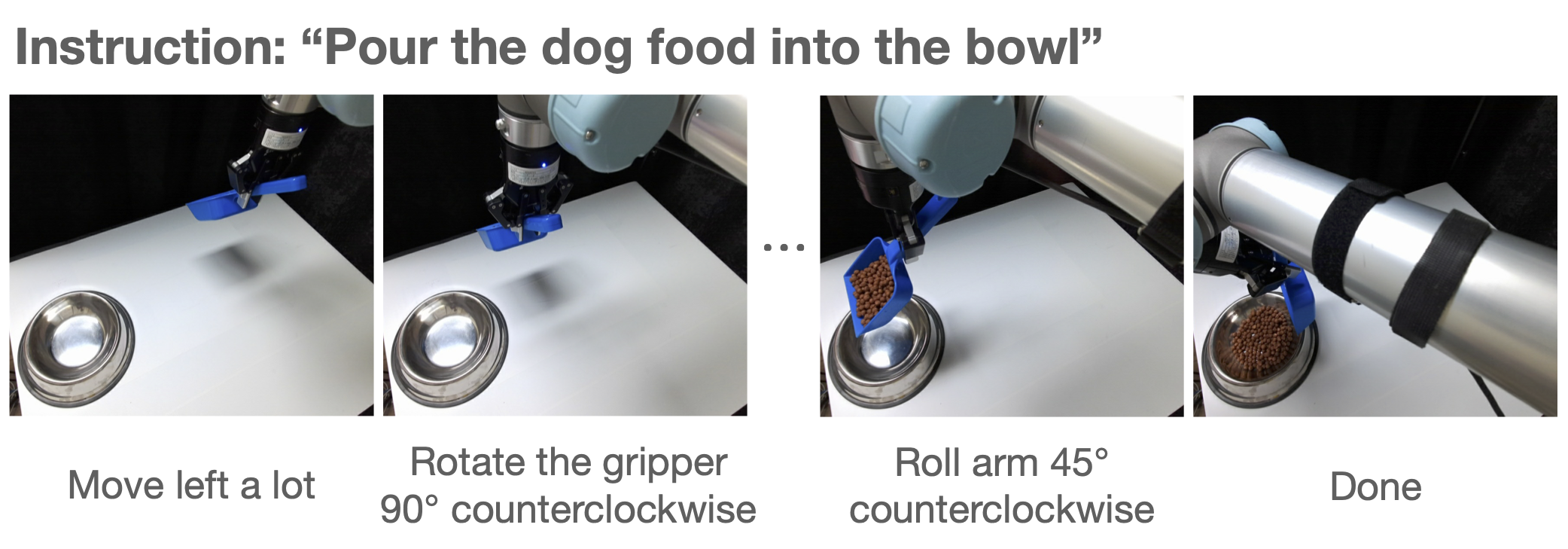}
\caption{Overview of language-guided teleoperation.}
\vspace*{-0.25cm}
\end{figure}

We argue that natural language is an intuitive and accessible interface for robot learning. We thus explore a method for training robotic skills through natural language. To this end, we propose a data collection framework that enables non-experts to collect in-domain robot data through natural language. It consists of two steps: language-based teleoperation and stochastic trajectory augmentation (STA). Figure~\hyperref[fig:teaser]{1} illustrates language-based teleoperation in which a human collects data for a skill described in the instruction (\textit{e.g.,} ``pour the dog food into the bowl''). The human first provides natural language supervision (\textit{e.g.,} ``move left a lot'') in each state. The large language model (LLM)~\cite{gpt4} then translates this supervision into appropriate robotic behavior, which is ultimately executed by the robot. By repeating this process, we obtain a collection of robot demonstrations, where each state transition is associated with corresponding language supervision. After the language-based teleoperation, STA augments the demonstration into alternative trajectories. Specifically, it stochastically drives the robot into novel states that were not explicitly covered in the original demonstrations. STA then automatically labels the appropriate behavior at these novel states using a simple heuristic. In other words, STA generates new trajectory data, expanding the diversity of the training dataset beyond the original demonstrations.

We introduce a vision-language-action (VLA) model that learns language-conditioned visuomotor policies from natural language supervision, which we call CLIP-RT (CLIP-based Robotics Transformer). A key idea is to leverage natural language as supervision to train visuomotor policies—inspired by CLIP~\citep{radford2021learning}, which uses language as a training signal for visual representation learning. CLIP-RT employs CLIP models trained in Internet-scale data~\cite{schuhmann2022laionb,fang2023data} and directly adapts them to predict language-based motion primitives (\textit{e.g.,} ``move the arm forward by 10cm'') through contrastive imitation learning. Specifically, our model learns to measure the pairwise similarity between language supervision and contextual information (\textit{i.e.,} current scene and language instruction) for language-conditioned policies. We train CLIP-RT through a two-step process: pretraining and in-domain fine-tuning. In the pretraining stage, we train our model on the large-scale robot learning dataset—Open X-Embodiment~\cite{padalkar2023open}—to improve generalization capabilities. The dataset does not contain language supervision, so we transform existing low-level robotic actions into templated natural language supervision to train CLIP-RT. During in-domain fine-tuning, CLIP-RT learns diverse robotic skills using our collected data.

Our contributions are fivefold. First, we propose CLIP-RT, a vision-language-action (VLA) model that learns language-conditioned policies from natural language supervision. Second, we propose a data collection framework that enables non-experts to collect robot data only through natural language and augment the human-collected demonstration data. Third, experiments demonstrate that CLIP-RT outperforms OpenVLA~\cite{kim24openvla} by 24\% in average success rates in 9 novel manipulation tasks. We further observe two important results: (1) language-based motion prediction and STA boost generalization capabilities of CLIP-RT and (2) CLIP-RT effectively learns shared structures across diverse robotic tasks, resulting in generalizable and transferable policies. Fourth, we demonstrate that CLIP-RT's language-based motion prediction capability enables collaboration with humans and large pretrained models~\citep{hurst2024gpt}, resulting in improved generalization. Fifth, to validate the generality of our method, we adapt CLIP-RT and evaluate it on the LIBERO simulation benchmark~\citep{liu2023libero} that includes offline, human-teleoperated demonstrations. CLIP-RT achieves strong results, an average success rate of 92.8\%, with an improved inference throughput of 163Hz.


%% file: sections/02_related_works.tex
\section{Related Work}

\noindent\textbf{Vision-Language-Action (VLA) Models.} Vision-language models (VLM) trained on Internet-scale data have been widely studied in robotics, including high-level planning~\cite{driess2023palme,hu2023look}, success detection~\cite{du2023vision}, and physical reasoning~\cite{gao2023physically}. In particular, previous work~\citep{brohan2023rt,padalkar2023open,belkhale2024rt,kim24openvla} directly fine-tunes VLMs to predict robotic actions. This category of models is called vision-language-action (VLA) models. CLIP-RT belongs to this category. Current VLA models discretize continuous action values (\textit{e.g.,} end-effector actions) into discrete action tokens and learn to generate a sequence of these tokens. Unlike existing VLA models, CLIP-RT is a \textit{discriminative} VLA model that predicts actions in a predefined list of actions, and these actions are represented in natural language (\textit{e.g.}, ``move the arm left'') rather than low-level control commands.

\noindent\textbf{Collecting Real-World Robot Demonstrations.} Data collection has become an increasingly important challenge in robot learning. Previous works have collected real-world robot demonstrations through various interfaces, such as teleoperation devices~\cite{fu2024mobile,abbeel2010autonomous}, virtual reality (VR)~\cite{zhang2018deep,seo2023deep}, and kinesthetic teaching~\cite{billard2006discriminative,maeda2017probabilistic,eteke2020reward}. Some studies introduce natural language interfaces~\cite{liu2023interactive,belkhale2024rt} for data collection, but they are often used in limited scenarios. RT-H~\cite{belkhale2024rt} and OLAF~\cite{liu2023interactive} first train visuomotor policies using data collected from other interfaces (\textit{e.g.,} VR). During deployment, humans provide language feedback to correct robotic behaviors, and policies are updated based on this feedback. In other words, these methods focus on refining learned policies for \textit{existing} skills. In contrast, our focus is to teach \textit{any desired} skills by collecting complete demonstration trajectories through language-based teleoperation. To achieve this, our framework uses the in-context learning capabilities of large language models (LLMs)~\cite{he2023annollm} to translate language supervision into action. \vspace*{0.1cm}

\noindent\textbf{Language-Conditioned Policies.} The research community has made extensive efforts to develop robotic systems that can follow language instructions~\cite{kollar2010toward,chen2011learning,thomason2020vision,shridhar2020alfred,kim2023gvcci,kang2024prograsp}, often training language-conditioned policies~\citep{stepputtis2020language,lynch2020language,shridhar2022cliport,jang2022bc,mees2022calvin,brohan2022rt,brohan2023rt,kim24openvla,belkhale2024rt}. We train language-conditioned visuomotor policies through imitation learning, similar to existing studies. Unlike existing studies, we train language-conditioned policies with contrastive imitation learning, which combines the ideas of contrastive learning~\cite{radford2021learning} with imitation learning~\cite{pomerleau1988alvinn} for more discriminative representations of robotic behaviors.

%% file: sections/03_approach.tex
\section{Approach}

\begin{figure*}[!t]
    \centering
    \label{fig:overview}
    \includegraphics[width=\textwidth]{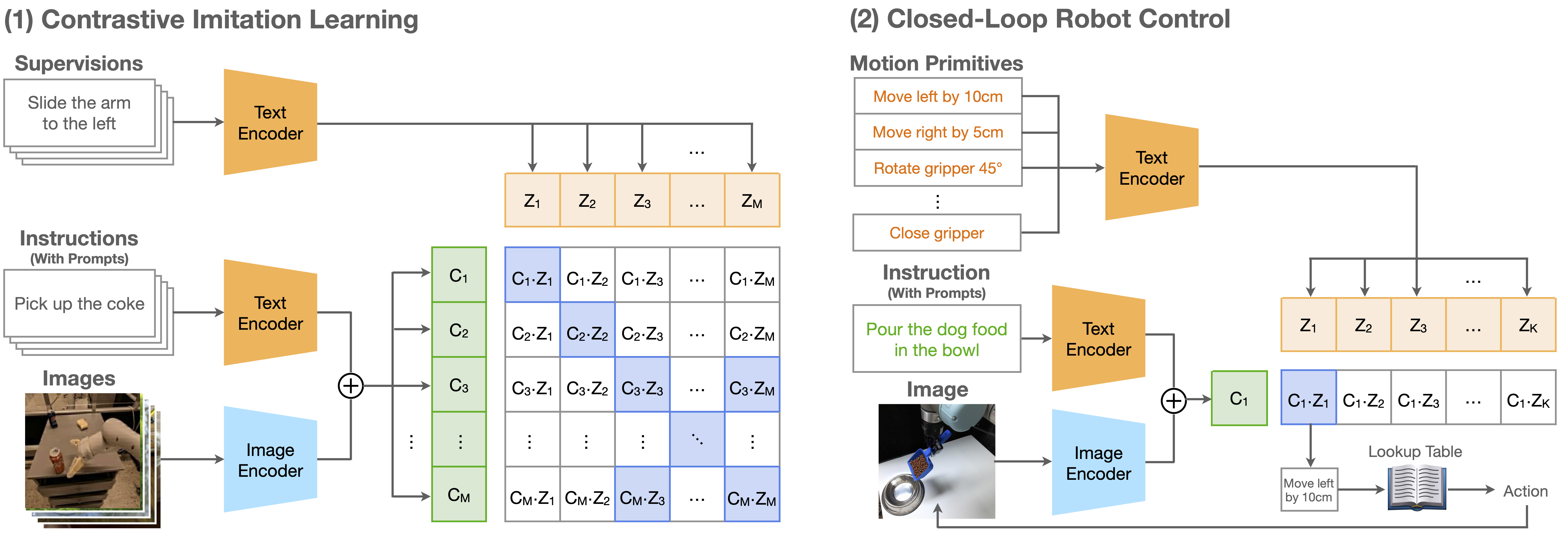}
    \caption{\textbf{Overview of CLIP-RT.} CLIP-RT learns to optimize the pairwise similarity between the context and natural language supervision through contrastive imitation learning. At test time, CLIP-RT predicts the language-based motion primitive with the highest similarity from a list of language motions. We append a simple text prompt to instructions: \textit{What motion should the robot arm perform to complete the instruction $\left\{\text{instruction}\right\}$?}}
\end{figure*}

\subsection{Preliminaries}

\noindent\textbf{Language-Conditioned Imitation Learning.} A robot dataset $\mathcal{D} = \{(\tau_n, \ell_n)\}_{n=1}^N$ consists of a demonstration trajectory $\tau$ paired with language instruction $\ell$. Each trajectory contains a sequence of visual observations and expert actions $\tau_n = \{(v_1, a_1), \ldots ,(v_{\vert \tau_n \vert}, a_{\vert \tau_n \vert})\}$. The goal of language-conditioned imitation learning is minimizing the negative log-likelihood of the expert action $a_t$ given the observation history $v_{1:t} = (v_1, \ldots, v_t)$ and language instruction $\ell$: 
\begin{equation}
    \mathcal{L}_{\mathrm{IL}} = - \mathbb{E}_{(\tau, \ell) \sim \mathcal{D}} \left[ \sum_{t=1}^{\vert \tau \vert} \mathrm{log} \, \pi_\theta(a_t | v_{1:t}, \ell) \right]
\end{equation}
where $\pi_\theta$ denotes the policy model with model parameters $\theta$. For vision-language action (VLA) models, $\theta$ is initialized from the parameters of vision-language models (VLMs). To maintain consistency with the pretraining setup of the VLMs, existing VLA models~\citep{brohan2023rt,kim24openvla,belkhale2024rt} typically use a single-image observation $v_t$ rather than utilizing the full observations $v_{1:t}$. At test time, the policy model performs closed-loop robot control until it completes language instructions. \vspace{0.2cm}

\noindent\textbf{Contrastive Language-Image Pretraining (CLIP)}~\citep{radford2021learning} is a method to learn visual representations from natural language supervision at scale. Using the contrastive objective, CLIP trains an image encoder $f(\cdot)$ and a text encoder $g(\cdot)$ on 400M image-text pairs. Given a mini-batch of $M$ image-text pairs $\left\{(I_i, T_i)\right\}_{i=1}^M$, the two encoders are jointly optimized to maximize the similarity between the correct pairs of image and text $(I_i, T_i)$ while minimizing the similarity for incorrect pairs $(I_i, T_{j \ne i})$. As we describe later, we modify the contrastive loss to make CLIP-RT learn language-conditioned policies.

\subsection{CLIP-Based Robotics Transformer (CLIP-RT)}

\noindent\textbf{Natural Language Supervision.} Inspired by CLIP~\cite{radford2021learning}, which uses natural language as a training signal, we built a model to learn robotic policies from natural language. We define natural language supervision as language-based guidance that directs a robot’s motion in specific states to complete given instructions. This typically involves shifting the robot's position, orientation, or gripper state (see Appendix~\hyperref[tab:nls]{A}). As we discuss later, each supervision is associated with a specific low-level action. Learning from natural language supervision offers several advantages. It establishes a clear hierarchy between initial instruction and language supervision, enabling models to learn \textit{shared structures} across diverse tasks~\cite{belkhale2024rt}. Furthermore, language-based learning fosters collaboration with language-capable entities like humans or other AI systems. \vspace*{0.2cm}

\noindent\textbf{Contrastive Imitation Learning (CIL).} We describe contrastive imitation learning in Figure~\hyperref[fig:overview]{2} (left). CLIP-RT takes a mini-batch of $M$ triplets $\left\{(v_i, \ell_i, u_i)\right\}_{i=1}^M$, where $v$, $\ell$, and $u$ denote image observation, instruction, and language supervision. CIL aims to optimize the pairwise similarities in the set $\left\{((v_i, \ell_i), u_j) | i, j \in  \left\{1, \ldots, M \right\}\right\}$. Specifically, CLIP-RT first extracts vector embeddings of $v_i$, $\ell_i$ and $u_j$ using the CLIP model's image encoder $f(\cdot)$ and the text encoder $g(\cdot)$, and subsequently combines the image and instruction embeddings:
\begin{equation}
    \mathbf{c}_i = f(v_i) + g(\ell_i), \;\;\; \mathbf{z}_j = g(u_j)
\end{equation}
where $\mathbf{c}_i$ represents the context that encapsulates the robot's current visual state and its explicit goal. $\mathbf{z}_j$ represents the immediate action that should be taken given the context. We design the loss function as:
\begin{equation}
\begin{split}
    \mathcal{L}_{\mathrm{CIL}} = - \frac{1}{M^2} \sum_{i=1}^M \sum_{j=1}^M & \Big[y_{ij} \log \sigma(\hat{\mathbf{c}}_i \cdot \hat{\mathbf{z}}_j) \\ &+ (1 - y_{ij}) \log (1 - \sigma(\hat{\mathbf{c}}_i \cdot \hat{\mathbf{z}}_j)) \Big]
\end{split}
\end{equation}
where $\hat{\mathbf{c}}_i = \frac{\mathbf{c}_i}{\lVert \mathbf{c}_i \rVert_2}$ and $\hat{\mathbf{z}}_j = \frac{\mathbf{z}_j}{\lVert \mathbf{z}_j \rVert_2}$ are normalized vector embeddings of $\mathbf{c}_i$ and $\mathbf{z}_j$. $\sigma(\cdot)$ is a sigmoid activation function and $y_{ij} \in \left\{0, 1\right\}$ denotes a label for pairwise similarity. The loss function maximizes the cosine similarity between context and language supervision for positive pairs, while minimizing it for negative pairs. The label $y_{ij}$ is basically one if $i=j$; otherwise, it is zero. In other words, $((v_i, \ell_i), u_{i})$ are positive pairs and $((v_i, \ell_i), u_{j \ne i})$ are negative pairs. However, the mini-batch often contains semantically interchangeable supervisions, such as ``move upwards'' and ``raise the arm''. Thus, CIL consults low-level actions $a_i$ associated with language supervision $u_i$ and treats the pair $((v_i, \ell_i), u_{j \ne i})$ as positive if two supervisions share the same low-level action. As a result, $y_{ij}$ is one if $i=j \text{ or } a_i = a_j$ (see the blue boxes in Figure~\hyperref[fig:overview]{2}); otherwise, it is zero. Consequently, CLIP-RT learns to measure the likelihood of each motion described in language, given visual observation and language instruction. \vspace*{0.2cm}   

\noindent\textbf{Pretraining.} We train CLIP-RT on the Open X-Embodiment (OXE) dataset~\cite{padalkar2023open}, which contains 2.4M robotic trajectories from 70 individual datasets. We specifically use the OXE data curated by \citet{kim24openvla} to train CLIP-RT. However, the data do not contain natural language supervision, so we extract language supervision from low-level action similar to recent studies~\cite{belkhale2024rt,zawalski2024robotic}. Specifically, the low-level action is represented as a 7-dimensional vector consisting of the end-effector's delta positions, delta orientations, and the gripper open/close. We identify the entry with the dominant value and its corresponding axis for each action. Based on this information, we transform low-level actions into one of 899 templated natural language supervisions (see Appendix~\hyperref[tab:nls]{A}). As a result, we train CLIP-RT on approximately 18.1M transition data through contrastive imitation learning. It requires four H100 GPUs for one day with a batch size of 128. \vspace*{0.2cm}

\noindent\textbf{In-Domain Fine-Tuning.} After pretraining, we fine-tune CLIP-RT on in-domain data via contrastive imitation learning. The in-domain dataset consists of 21K transitions in 18 robotic manipulation tasks, collected through our data collection framework. Details about the dataset and data collection are discussed in the following sections (\ref{ssec:robot_data_collection} and~\ref{ssec:task_and_dataset}). \vspace*{0.2cm}

\noindent\textbf{Closed-Loop Robot Control.} Figure~\hyperref[fig:overview]{2} (right) shows an overview of closed-loop robot control. At each time step, CLIP-RT computes pairwise similarities between the context and a list of language-based motion primitives. Our model selects the motion with the highest probability. This selected motion is translated into a lower-level end-effector action based on a predefined lookup table (see Appendix~\hyperref[tab:lookup]{B}). Finally, the translated end-effector action is executed using inverse kinematics (IK). Unlike existing Transformer-based policy models~\citep{brohan2023rt, brohan2022rt,padalkar2023open,belkhale2024rt,kim24openvla} relying on autoregressive decoding, CLIP-RT predicts each action in a \textit{single} forward pass since it is a discriminative model. CLIP-RT runs at 16Hz on one H100 GPU and 8Hz on one NVIDIA RTX 3090 GPU, both using float32 precision. These results are achieved without any speed-up tricks (\textit{e.g.,} model quantization). Details regarding frequencies are discussed in Appendix~\ref{ssec:runtime}. \vspace*{0.2cm}

\noindent\textbf{VLM Backbone \& Codebase.} CLIP-RT maintains the original CLIP model architecture without any new parameters. As our backbone model, we employ ViT-H-14-378-quickgelu~\cite{fang2023data,ilharco_gabriel_2021_5143773}, an open-source CLIP model of 986M ($\approx$1B) parameters that achieves state-of-the-art performance in zero-shot image classification~\cite{imagenet15russakovsky} at the time of writing. It consists of an image encoder~\cite{dosovitskiy2020image} and a text encoder~\cite{radford2019language}, both built on Transformer~\cite{vaswani2017attention}. All model configurations can be found in the OpenCLIP codebase~\cite{ilharco_gabriel_2021_5143773}. A key advantage of this codebase is that strong CLIP models are continuously updated to the dashboard, enabling users to easily use them through a plug-and-play approach.


\subsection{In-Domain Data Collection}

\noindent\textbf{Language-Based Teleoperation.} \label{ssec:robot_data_collection} This step aims to collect a few robot demonstrations for each skill only through natural language. To this end, we employ a large language model (LLM)~\cite{gpt4} and design a scenario where users collect in-domain data through interactions with the LLM. Specifically, users first provide an initial language instruction for each skill. Then, they provide natural language supervision in each state to complete the instruction. The LLM translates the language supervision into the low-level end-effector action based on a detailed text prompt (see Appendix~\hyperref[tab:prompt]{C}). Finally, the camera captures the current image observation and the robot executes the translated action. Consequently, we can obtain a sequence of tuples $\left\{(v_i, \ell_i, u_i, a_i) \right\}_{i=1}^N$ containing visual observation, instruction, natural language supervision, and low-level action. We collect 10 episodes for each skill through this process.
\begin{figure}[!t]
    \centering
    \includegraphics[width=\columnwidth]{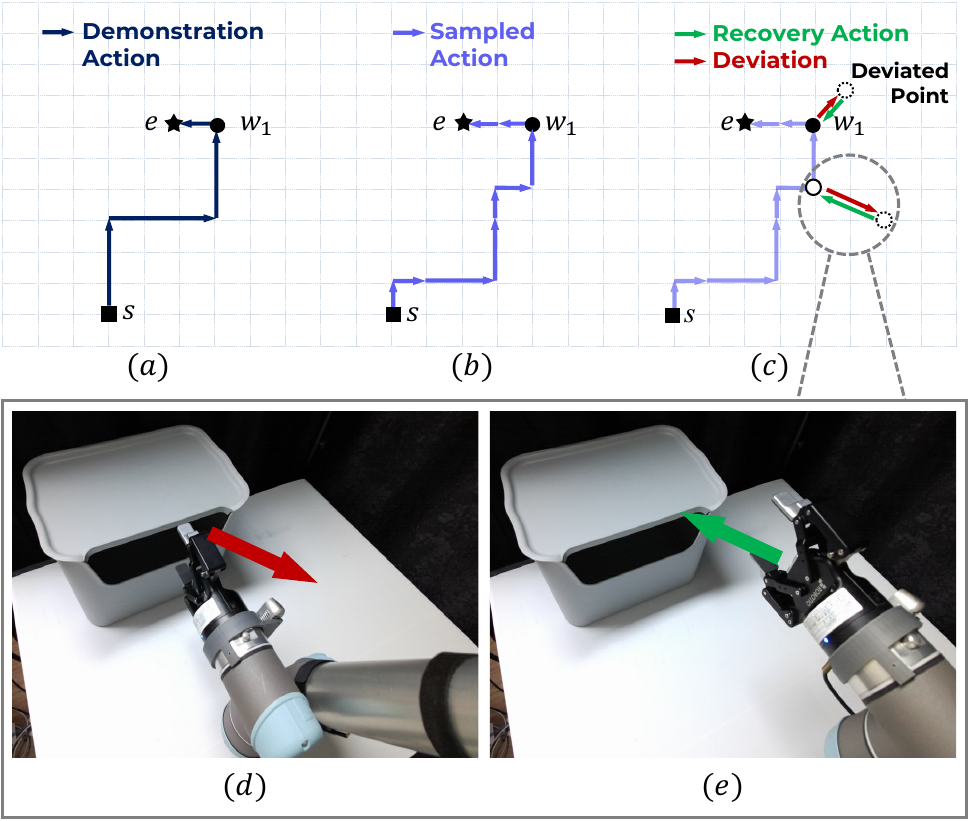}
    \vspace*{-0.2cm}
    \caption{\textbf{A simplified 2D example of stochastic trajectory augmentation (STA).}  
    (a): a demonstration trajectory from the start \(s\) to the endpoint \(e\), passing through a waypoint \(w_1\).  
    (b): a sampled trajectory generated by the diversification phase. 
    (c)-(e): a visualization of the recovery phase. 
    }
    \label{fig:std}
\end{figure}

\noindent\textbf{Stochastic Trajectory Augmentation (STA)} aims to augment the demonstration data collected from language-based teleoperation. Before delving into the details, we first define a \emph{waypoint} as a key state in demonstrations that satisfies either of the following conditions: (1) the gripper state changes (\textit{i.e.,} open $\rightarrow$ close or close $\rightarrow$ open) or (2) the cumulative progress of delta positions along any axis reverses. For example, $w_1$ in Figure~\ref{fig:std}-(a) is a waypoint since cumulative progress on a horizontal axis starts to reverse at $w_1$. STA consists of two phases: \emph{diversification phase} and \emph{recovery phase}. The diversification phase first builds alternative trajectories toward each waypoint (see Figure~\ref{fig:std}-(b)) by sampling a new action sequence. The robot then executes each action in the sequence, recording an image in every state it visits. In the recovery phase, STA drives the robot into novel states that deviate from the planned trajectory (see Figure~\ref{fig:std}-(d)) and then executes a recovery action, a simple reversal of the deviation to return to the trajectory (see Figure~\ref{fig:std}-(e)). Note that STA records only the recovery actions and images in the deviated states, not the deviation data. By alternating these two phases, STA automatically expands the diversity of the original demonstrations, potentially improving the robustness of policies under varied states. Further details of STA are discussed in Appendix~\ref{sec:appendix_std_algorithm}.

\begin{figure*}[!t]
    \centering
    \includegraphics[width=\textwidth]{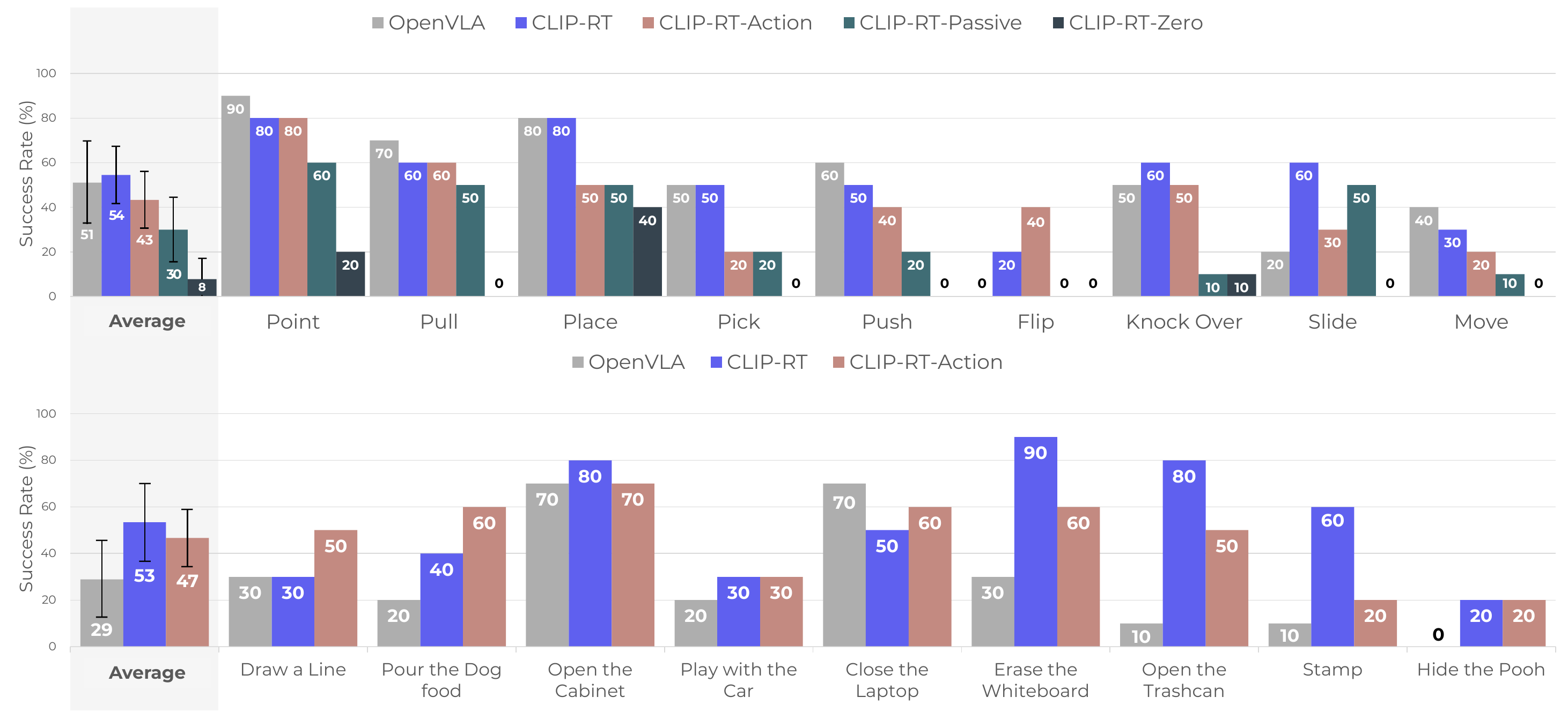}
    \caption{\textbf{Success rates on 9 Common tasks (top) and 9 Novel tasks (bottom).} We conduct experiments using all compared methods on Common tasks and three models (CLIP-RT, OpenVLA and CLIP-RT-Action) on Novel Tasks. The success rate for each task is measured by averaging the results of ten trials. Average success rates of all tasks are shown on the left for both Common and Novel task sets. Tasks are arranged from left to right based on their average number of steps per episode in the training data. The task on the right indicates that it requires more steps in average compared with the task on the left.}
    \label{fig:result_performance}
\end{figure*}

%% file: sections/04_experiment.tex
{\section{Experiments on Real-World Robotic Manipulation}}

\subsection{Tasks \& Dataset}
\label{ssec:task_and_dataset}
We train and evaluate our models in 18 robotic manipulation tasks, categorized into two groups: \textit{Common} and \textit{Novel}. 
\textbf{Common tasks} consist of nine tasks closely aligned with those in the Open X-Embodiment dataset~\citep{padalkar2023open}. These tasks include common manipulation skills, such as “\textit{pick the \texttt{<obj>}}” and “\textit{place the \texttt{<obj>} on the \texttt{<obj>}}”.
In contrast, \textbf{Novel tasks} include nine tasks barely observed during pretraining on the Open X-Embodiment dataset, such as “\textit{stamp on \texttt{<obj>}}”, “\textit{play with the toy car}”, and “\textit{erase the whiteboard}”. This set of tasks serves as a benchmark for evaluating the model's ability to acquire new skills using in-domain data. We first collect in-domain data through language-based teleoperation, gathering 10 episodes per task, resulting in 911 transitions for Common tasks and 1,123 transitions for Novel tasks. Leveraging stochastic trajectory augmentation (STA), we augment each demonstration with 3 additional trajectories across all tasks. This augmentation increases the dataset size to approximately 11K transitions for Common tasks and 10K transitions for Novel tasks. Unless stated otherwise, all the models compared were trained on the same dataset. We provide details of each task, along with visualizations, in Appendix~\ref{sec:appendix_dataset}.

\subsection{Robotic Platform}
We perform experiments using a physical robot arm, 6-DoF Universal Robots (UR5) with a two-finger gripper. We provide more details about the robotic platform in the Appendix~\ref{sec:robot_platform}.

\subsection{Experiments on Common and Novel Tasks}
\label{ssec:exp_multitask}
We train and evaluate CLIP-RT on both Common and Novel tasks, comparing with diverse baselines. We introduce baseline models and then discuss the results in detail.

\noindent\textbf{Baselines.} We compare CLIP-RT with four methods, including the state-of-the-art method and ablated versions of our model:
\begin{itemize}
  \item \textbf{CLIP-RT} is our proposed model, pretrained on the Open X-Embodiment (OXE) dataset~\citep{padalkar2023open} and further fine-tuned using our in-domain data.
  \item \textbf{OpenVLA}~\citep{kim24openvla} is a state-of-the-art, open-source vision-language-action (VLA) model. This model leverages the 7B-parameter Llama2 language model~\citep{touvron2023llama} and a visual encoder that combines pretrained features from DINOv2~\citep{oquab2023dinov2} and SigLIP~\citep{zhai2023sigmoid}. We also fine-tune OpenVLA on the same in-domain data as CLIP-RT by using low-level 7D end-effector actions as supervision.
  \item \textbf{CLIP-RT-Action} is a variant of CLIP-RT where each motion is mapped to existing text tokens that are not frequently used in the vocabulary, similar to existing VLA models~\citep{kim24openvla,brohan2023rt,brohan2022rt,padalkar2023open}. In other words, CLIP-RT-Action represents actions as learned action tokens, rather than representing in natural language. It is also pretrained on the OXE dataset and fine-tuned on in-domain data.
  \item \textbf{CLIP-RT-Passive} is another ablated model of CLIP-RT, which excludes data collected from stochastic trajectory augmentation (STA) and relies solely on data from language-based teleoperation.
  \item \textbf{CLIP-RT-Zero} is an ablated model trained solely on the OXE dataset without accessing any in-domain data.
\end{itemize}
\vspace*{0.2cm}

\noindent\textbf{Results on Common Tasks.} We compare CLIP-RT with all baseline models on Common tasks. The results are summarized in the upper row of Figure~\hyperref[fig:result_performance]{4}. CLIP-RT achieves an average success rate of 54\%, outperforming all baselines, including OpenVLA and three ablative models. {While CLIP-RT outperforms OpenVLA on average, OpenVLA still shows better performance on four basic tasks—Point, Pull, Push, and Move.} When comparing CLIP-RT with CLIP-RT-Action, we observe that the use of natural language supervision significantly increases performance on Common tasks (43\% $\rightarrow$ 54\%). We hypothesize that CLIP-RT effectively leverages the rich vision-language representations of the pretrained CLIP model~\cite{radford2021learning}, allowing it to align language-based motions with semantic concepts. Furthermore, CLIP-RT-Passive, which omits stochastic trajectory augmentation (STA), struggles in most tasks, highlighting the critical role of STA in performance. This suggests that STA enhances robustness and generalization, enabling CLIP-RT to adapt to novel situations. We refer readers to Appendix~\ref{ssec:qualitative_std} for a more detailed analysis on the effect of STA. Finally, CLIP-RT-Zero, despite being trained in the large-scale robot learning dataset~\cite{padalkar2023open}, shows 8\% on average success rates, underscoring the need for in-domain fine-tuning. \vspace*{0.2cm}

\noindent\textbf{Results on Novel Tasks.} We compare CLIP-RT with OpenVLA and CLIP-RT-Action on 9 Novel tasks. In the lower row of Figure~\hyperref[fig:result_performance]{4}, CLIP-RT achieves an average success rate of 53\%, outperforming these baselines. Notably, CLIP-RT maintains its average success rates on Novel tasks compared to those of Common tasks, but we observe a significant performance drop of OpenVLA on Novel tasks (51\% $\rightarrow$ 29\%). These findings suggest that CLIP-RT generalizes more effectively to tasks that are barely observed in the pretraining dataset. {To verify the statistical significance of the performance difference between CLIP-RT and OpenVLA, we conduct a t-test. The resulting p-value is $p = 1.74 \times 10^{-9}$, indicating that CLIP-RT significantly outperforms OpenVLA.} 

\begin{figure}[!t]
    \centering
    \includegraphics[width=\columnwidth]{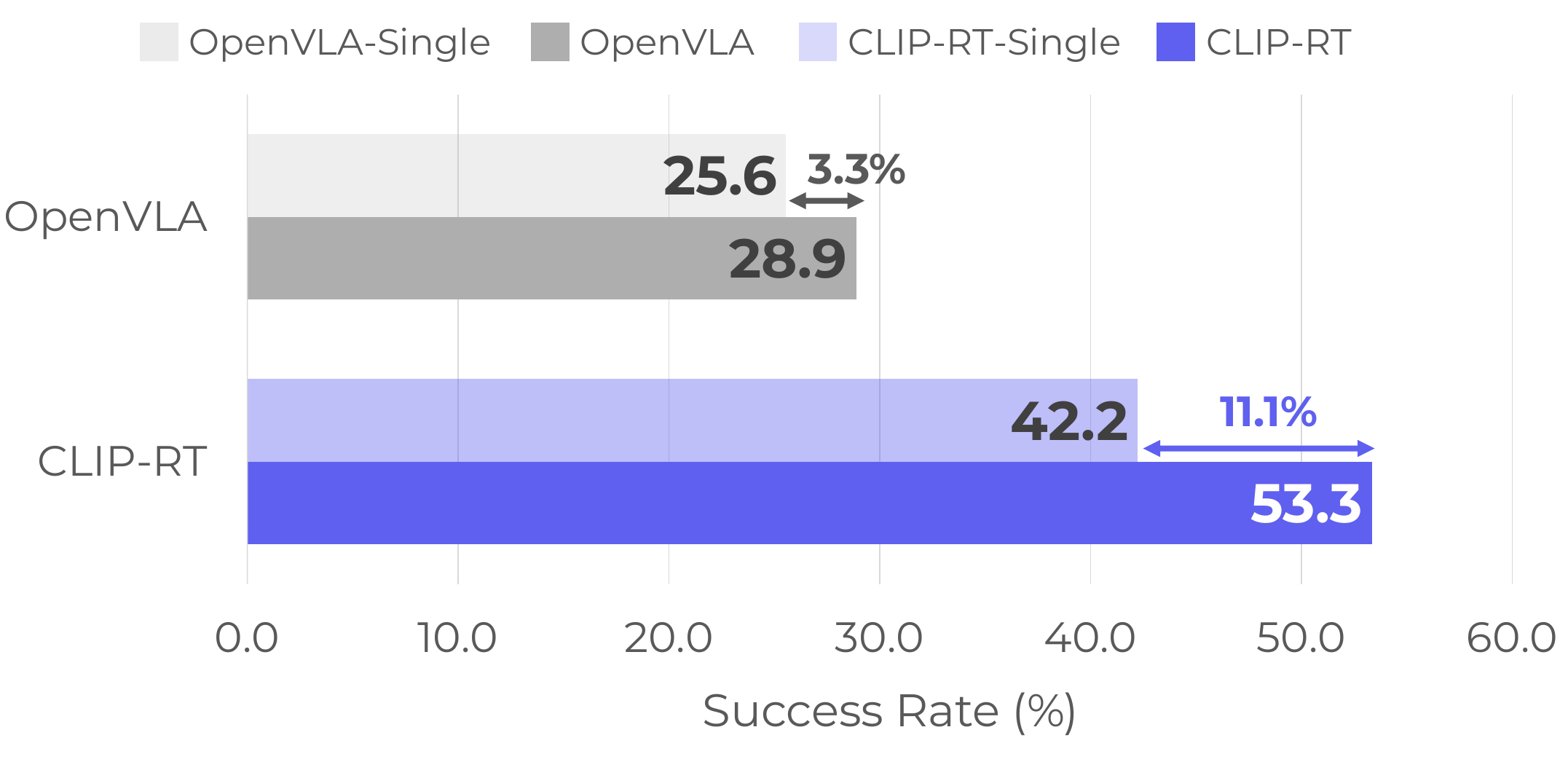}
    \caption{
    \textbf{A comparison of multi-task and single-task policies on Novel tasks.} The performance of each task is in Figure~\hyperref[fig:novel_single]{12} of Appendix.
    }
    \label{fig:single}
    \vspace*{-0.5cm}
\end{figure}

\subsection{In-Depth Analysis of Generalization}

We investigate the source of CLIP-RT's improved generalization on Novel tasks. We conduct analyses along three axes: (1) a comparison between multi-task and single-task policies, (2) the effect of natural language supervision, and (3) few-shot generalization.

\noindent\textbf{Comparison Between Multi-Task and Single-Task Policies.} Where does the significant performance gap between CLIP-RT and OpenVLA on Novel tasks come from? One of our hypotheses is that CLIP-RT effectively learns the \emph{shared structure} across diverse robotic tasks by utilizing language-based motion primitives as basic building blocks. To verify this, we train a single-task policy for each Novel task and evaluate the performance of each model. In other words, 9 individual single-task policies for both CLIP-RT and OpenVLA are evaluated. The results are summarized in Figure~\hyperref[fig:single]{5}. OpenVLA-Single and CLIP-RT-Single denote the performance of single-task policies for each model. Compared to multi-task policies, both models show performance drops with single-task policies—3.3\% drop for OpenVLA and 11.1\% drop for CLIP-RT. This suggests that multi-task policy learning benefits both models, but CLIP-RT, with its larger performance gap, benefits more from shared knowledge across tasks. This highlights that CLIP-RT facilitates the learning of more generalizable and transferable policies compared with OpenVLA. \vspace*{0.1cm}

\begin{figure}[!t]
    \centering
    \includegraphics[width=1\columnwidth]{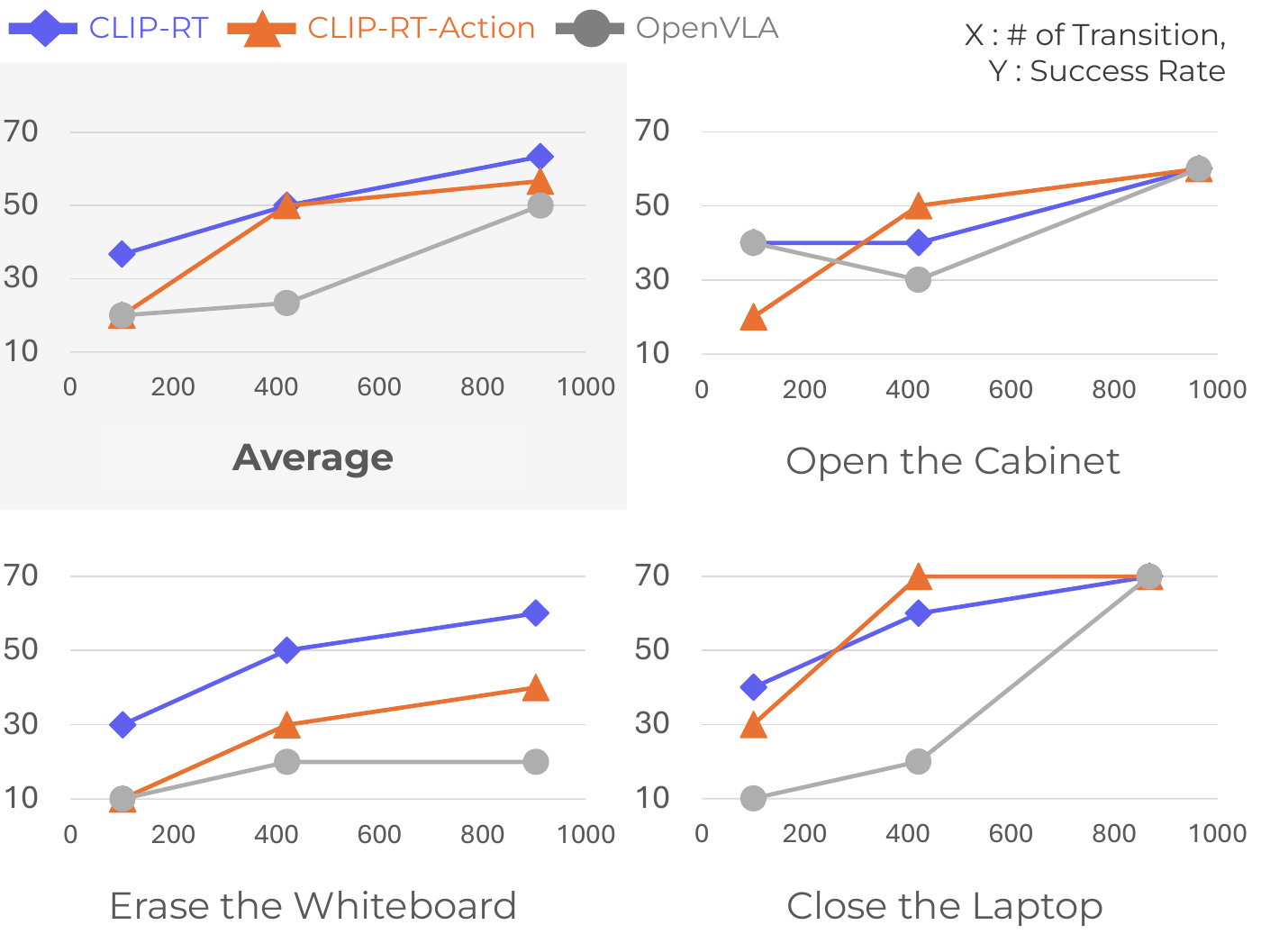}

    \caption{
        \textbf{Results on few-shot learning.} 
        We report the performance of CLIP-RT, CLIP-RT-Action, and OpenVLA with 1, 5, and 10 demonstrations (from left to right in each graph). The x-axis denotes the number of transitions actuall provided, and the y-axis indicates the task success rate.
    }
    \label{fig:fewshot}
\end{figure}

\noindent\textbf{Effect of Natural Language Supervision.} In Figure~\hyperref[fig:result_performance]{4}, CLIP-RT outperforms CLIP-RT-Action on both Novel and Common tasks. This indicates that the use of natural language supervision also enhances CLIP-RT's generalization capabilities. We visualize the action embeddings of both models to further analyze the impact of natural language supervision in Appendix~\ref{sec:embedding_vis}.

\noindent\textbf{Few-Shot Generalization.} Does CLIP-RT perform effectively with a limited amount of in-domain data? We further investigate this by evaluating learned policies, assuming fewer demonstrations (\textit{i.e.,} 1, 5, and 10) are provided. Specifically, we compare CLIP-RT with OpenVLA and CLIP-RT-Action on three Novel tasks, where all models performs relatively well on average. As shown in Figure~\ref{fig:fewshot}, CLIP-RT demonstrates improved performance in few-shot policy learning, especially in the single demonstration setting. Such few-shot adaptation is particularly crucial for robotics, where pretraining data (\textit{e.g.,} \citet{padalkar2023open}) cannot cover all real-world tasks, necessitating models that can rapidly acquire new skills from minimal demonstrations.

\input{sections/04.3_collaboration}

%% file: sections/04.3_collaboration.tex
\subsection{Collaborative Capabilities of CLIP-RT}
\label{ssec:collaboration}
Learning and reasoning about actions in natural language offer an additional benefit: collaborative problem-solving with language-capable entities. In this subsection, we explore how CLIP-RT collaborates with (1) humans by incorporating corrections and (2) large pretrained models via action refinement.

\begin{figure}[!t]
    \centering
    \includegraphics[width=0.85\columnwidth]{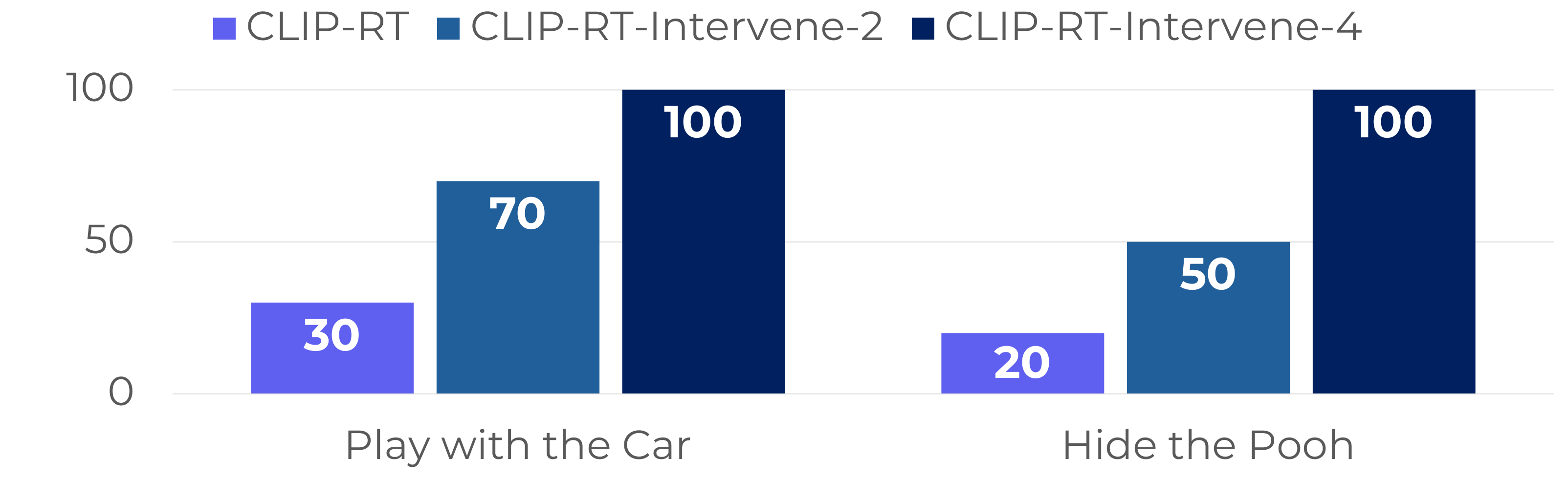}
    \caption{ 
        \textbf{Performance on varying numbers of human interventions.} Success rates of two challenging tasks under 0, 2, and 4 human corrections. Each success rate is measured by averaging the results of ten trials.
        }
    \label{fig:intervention}
\end{figure}

\noindent\textbf{Collaboration with Humans.} When CLIP-RT predicts an incorrect motion, humans can easily interpret the predictions and provide a correct motion in a certain state (e.g., ``rotate gripper 90 degrees''). We study two tasks in which CLIP-RT achieves its lowest success rates---\emph{Play with the Car} and \emph{Hide the Pooh}---and measure how a small number of human interventions affects performance. We set a maximum limit on the number of corrections per episode humans can provide: 2 and 4. Figure~\ref{fig:intervention} shows the task success rate with varying numbers of human interventions (0, 2, and 4). 
Without intervention, CLIP-RT's success rates are 30\% and 20\% on these two tasks. With two interventions, these rates increase to 70\% and 50\%, and with four interventions, both tasks achieve a 100\% success rate. These results demonstrate that even a few human corrections substantially improve CLIP-RT’s performance in challenging tasks. Since actions are expressed in language, humans can easily intervene with language corrections \vspace*{0.2cm}

\begin{figure}[!t]
    \centering
    \includegraphics[width=\columnwidth]{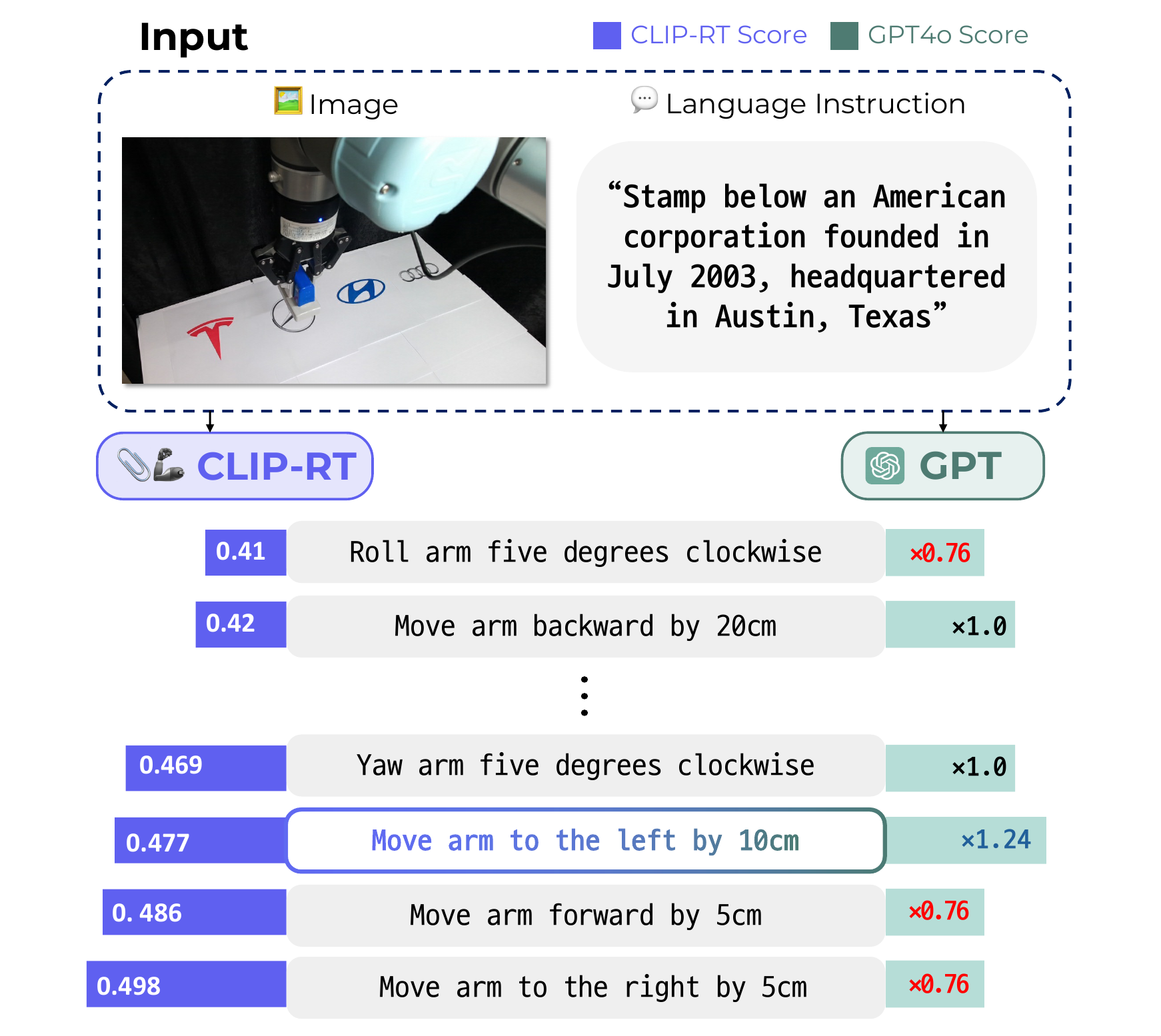}
        \caption{\textbf{Ensembling CLIP-RT and GPT outputs.}
        Given an image and language instruction (top), CLIP-RT produces initial scores for candidate actions (left). GPT then supplies multiplicative appropriateness factors for each action (right), which are applied to the CLIP-RT scores to determine the final action, "\textit{Move the arm to the left by 1cm}".}
    \label{fig:ensemble}
\end{figure}

\noindent \textbf{Collaboration with Large Pretrained Models.}
We also investigate how CLIP-RT can collaborate with a large pretrained model---GPT-4o~\cite{hurst2024gpt} (GPT for short)---through action refinement. 
As shown in Figure~\ref{fig:ensemble}, at each transition, we provide the current image observation and instruction to GPT. GPT then proposes a set of action candidates and labels them as either ``appropriate'' or ``inappropriate''. 
CLIP-RT incorporates this feedback by boosting the scores of actions deemed appropriate and penalizing those labeled as inappropriate.
In the example from Figure~\ref{fig:ensemble}, CLIP-RT initially assigns a high score to ``Move arm to the right'', but GPT labels this motion as inappropriate and provide positive rewards to the motion, ``Move arm to the left'', leading to a correct prediction. This GPT-guided approach broadens the range of instructions that CLIP-RT can handle, enabling it to execute instructions that require commonsense knowledge or high-level reasoning. For instance, CLIP-RT can benefit from collaboration with large pretrained models when given instructions like ``Stamp below an American corporation founded in July 2003, headquartered in Austin, Texas,'' as shown in Figure~\ref{fig:ensemble}. We provide several qualitative examples in Appendix~\ref{ssec:appendix_ensemble} to illustrate how large pretrained models can help perform out-of-distribution instructions requiring commonsense knowledge or complex reasoning. Furthermore, Appendix~\ref{ssec:prompt_gpt} discusses details about the GPT's text prompt and how exactly GPT’s decisions are integrated to CLIP-RT’s scores.

\subsection{Analysis on Failure Cases}

\begin{figure}[!t]
    \centering
    \includegraphics[width=\columnwidth]{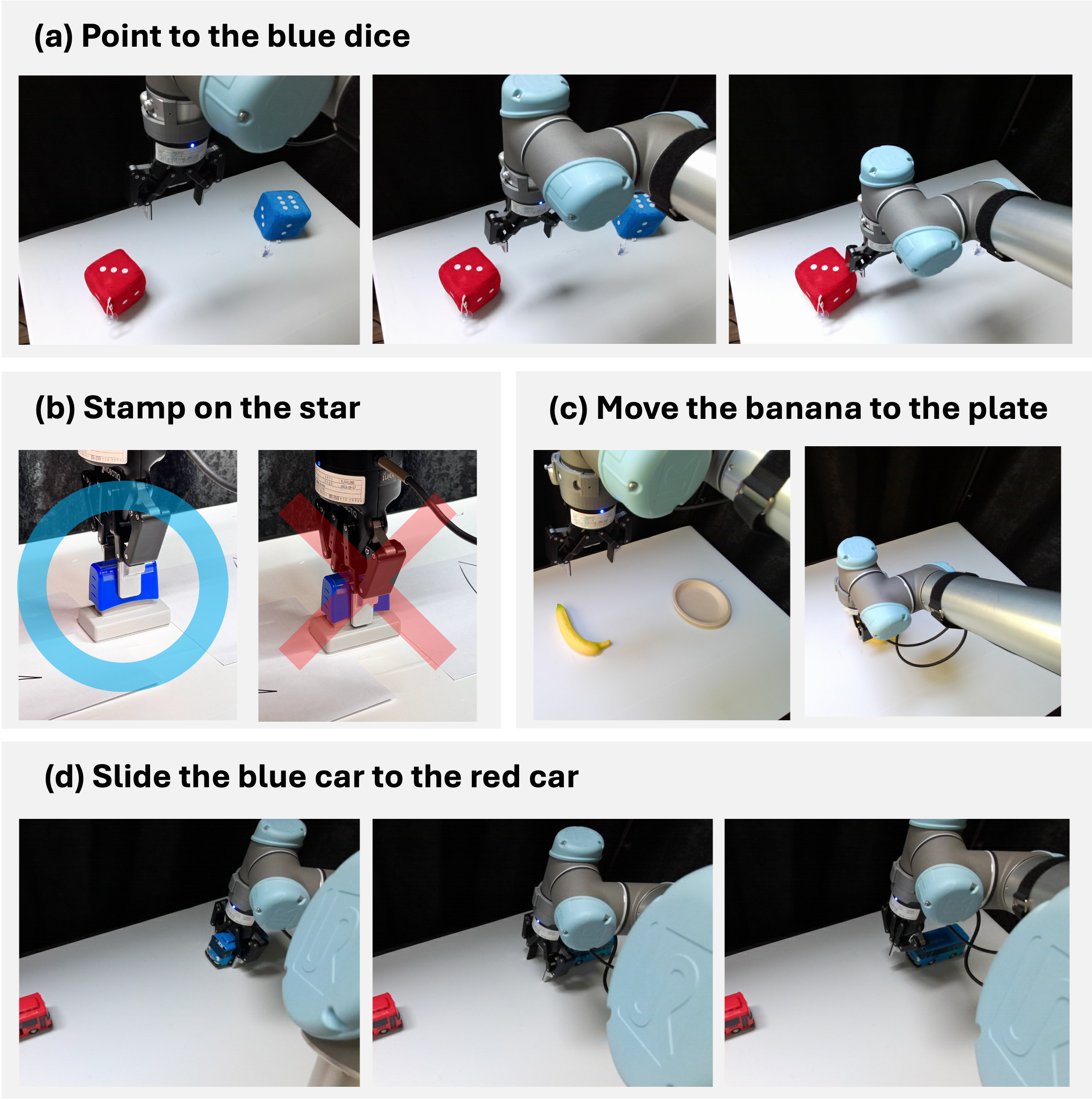}
    \caption{
    \textbf{Example failure cases of CLIP-RT.} 
    (a) CLIP-RT incorrectly identifies the target, pointing at the red dice instead of the blue dice. 
    It is difficult to detect the correct spatial relationship between the cup and the hanger based on the initial visual input. 
    (b) Failure in executing the ``Stamp on the star''. The left figure demonstrates a correct grasp of the stamp, whereas the right figure illustrates an incorrect grip that prevents successful task completion.
    (c) The robot arm completely obstructs the objects of interest, preventing accurate perception and manipulation.
    (d) The robot slips while attempting to slide the blue car and fails to recover by reopening the gripper and attempting to re-grasp the object.
    }
    \vspace*{-0.2cm}
    \label{fig:failure_cases}
\end{figure}

We visualize four types of failure cases. First, CLIP-RT has occasionally failed to comprehend the attributes of objects specified in instructions. For example, Figure~\ref{fig:failure_cases}-(a) depicts a scenario in which CLIP-RT is instructed to \emph{point to the blue dice}, but mistakenly pointed at the red dice instead. This examples confirms a need of more precise visual grounding. 

Second, CLIP-RT sometimes fail to execute tasks that require fine-grained control, such as \emph{Stamp on} \texttt{<obj>}. Figure~\ref{fig:failure_cases}-(b) illustrates an example of such a task. Based on the image observed from the current distance, it may be difficult to precisely determine whether the z-axis of the gripper is properly aligned to stamp on \texttt{<obj>}. This limitation is likely due to the reliance on 2D image inputs, which makes it challenging to accurately infer the 3D spatial information necessary for precise manipulation. The models, pretrained on large-scale image-text datasets, may not capture the depth and spatial nuances required for such tasks. Utilizing inputs like RGB-D images or point clouds might alleviate this issue.
Third, relying on images from a single viewpoint can lead to occlusions, as visualized in Figure~\ref{fig:failure_cases}-(c), particularly when the robot's arm obstructs the object of interest. Employing multiple camera angles could alleviate this issue by providing a more comprehensive view of the scene.

Fourth, stochastic trajectory augmentation (STA) relies on heuristic algorithms that may not capture the full diversity of possible trajectories. This is particularly evident in scenarios requiring recovery from failure states, such as when an object slips from the gripper, as shown in Figure~\ref{fig:failure_cases}-(d). The heuristics does not adequately represent the multitude of ways a robot might recover or adapt in these situations, potentially hindering the model's ability to generalize to unforeseen circumstances.


%% file: sections/05_libero.tex
\section{Experiments: Adapting CLIP-RT to Simulated Environments}

\setlength{\tabcolsep}{9pt}

\begin{table*}[t!]
\centering
\resizebox{\textwidth}{!}{
{
\begin{tabular}{lcccccccc}
\toprule
\multirow{6}{*}{Model} & \multirow{6}{*}{Size} & \multicolumn{2}{c}{Inference Efficiency} & \multicolumn{5}{c}{LIBERO Task Success Rates} \\
\cmidrule(lr){3-4} \cmidrule(lr){5-9}
& & \makecell{Throughput$\uparrow$ \\ (Hz)} & \makecell{Latency$\downarrow$ \\ (Sec)} & \makecell{Spatial$\uparrow$ \\ (\%)} & \makecell{Object$\uparrow$ \\ (\%)} & \makecell{Goal$\uparrow$ \\ (\%)} & \makecell{Long$\uparrow$ \\ (\%)} & \makecell{Average$\uparrow$ \\ (\%)} \\
\midrule
Octo~\citep{octo_2023} & 93M & - & - & 78.9 & 85.7 & 84.6 & 51.1 & 75.1 \\
DP (scratch)~\citep{chi2023diffusion} & 157M & - & - & 78.3 & 92.5 & 68.3 & 50.5 & 72.4 \\
Dita~\citep{hou2025dita} & 334M & - & - & 84.2 & 96.3 & 85.4 & 63.8 & 82.4 \\
OpenVLA~\citep{kim24openvla} & 7.5B & 4.2 & 0.240 & 84.7 & 88.4 & 79.2 & 53.7 & 76.5 \\
OpenVLA-OFT~\citep{kim2025fine} & 7.7B & 109.7 & 0.073 & \textbf{96.2} & \underline{98.3} & \textbf{96.2} & \textbf{90.7} & \textbf{95.3} \\
\midrule
\rowcolor{yellow!30}
\textbf{CLIP-RT+ (ours)} & 1.3B & \textbf{163.8} & \textbf{0.049} & \underline{95.2} & \textbf{99.2} & \underline{94.2} & \underline{83.8} & \underline{93.1} \\
\bottomrule
\end{tabular}
}
}
\caption{{\textbf{LIBERO task performance and inference efficiency results.} All models, except Diffusion Policy (DP)~\citep{chi2023diffusion}, were fine-tuned. Boldface scores represent the highest score, while underlined scores indicate the runner‑up.}}
\label{tab:libero}
\end{table*}

\setlength{\tabcolsep}{6pt}

While our primary focus is on training real-world robots through language-guided data collection, we further evaluate CLIP-RT on the LIBERO simulation benchmark~\citep{liu2023libero} to study the following questions:
\begin{itemize}
\item \textbf{Generality}: Is CLIP-RT applicable to environments with offline, human-teleoperated demonstration data?
\item \textbf{Performance}: Does CLIP-RT remain effective in a controlled simulation setting?

\end{itemize}

\begin{figure}[!t]
    \centering
    \includegraphics[width=0.8\columnwidth]{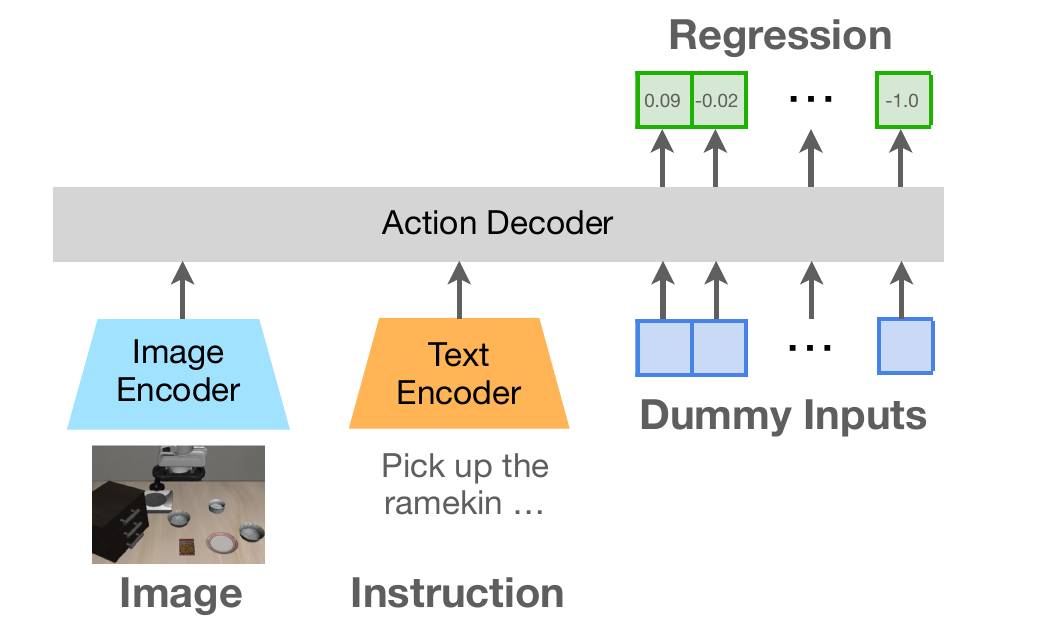}
        \caption{\textbf{Overview of CLIP-RT+ for LIBERO.}}
    \label{fig:reg}
\end{figure}

\subsection{Tasks \& Dataset}

We evaluate on four task suites of the LIBERO benchmark: LIBERO-Spatial, LIBERO-Object, LIBERO-Goal, and LIBERO-Long. These task suites assess policy generalization to diverse spatial relationships, objects, task goals, and long-horizon tasks. Each task suite contains 500 human-teleoperated demonstration data for 10 different tasks. By following the experimental setup in existing studies~\citep{kim2025fine,kim24openvla,hou2025dita}, we train and evaluate CLIP-RT on each task suite individually.

\subsection{Adapting CLIP-RT to the LIBERO Benchmark}
\label{ssec:libero_main}
Before describing how we adapt CLIP-RT to the LIBERO simulation benchmark, we acknowledge the inherent difficulty of directly representing the fine-grained, continuous human-teleoperated actions in LIBERO using natural language at a comparable level of abstraction. This discrepancy in abstraction levels necessitates the design of an alternative model architecture to enable effective action prediction in this setting. Accordingly, we simply add a 0.3B-parameter action decoder to the original CLIP-RT model to predict continuous actions. We refer to this model as CLIP-RT+. By following~\citet{kim2025fine}, we employ action chunking and parallel decoding. As shown in Figure~\ref{fig:reg}, the action decoder takes the image and instruction embeddings vectors from CLIP-RT and zero-valued empty tokens as inputs. We use the L1 regression-based objective to optimize the model. The action decoder shares the same model architecture with the CLIP-RT's text encoder. As a result, CLIP-RT+ is a 1.3B-parameter model. The size of the action chunk is 8, and the dimension of each action is 7. We train CLIP-RT+ using 8 NVIDIA H100 GPUs for 128 epochs with a batch size of 256.

\subsection{Results \& Discussions}

We compare CLIP-RT+ with the state-of-the-art models on the LIBERO simulation benchmarks, including OpenVLA~\citep{kim24openvla}, OpenVLA-OFT~\citep{kim2025fine}, Dita~\citep{hou2025dita}, DP~\citep{chi2023diffusion}, and Octo~\citep{octo_2023}. As shown in Table~\ref{tab:libero}, the recent state-of-the-art VLA model, OpenVLA-OFT~\citep{kim2025fine}, achieves the highest average success rate of 95.3\%. However, CLIP-RT+ shows comparable performance across all task suites with an average score of 93.1\%, while using 6x fewer parameters (1.3B) compared with OpenVLA-OFT (7.7B). Surprisingly, CLIP-RT+ attains a near perfect success rate (99.2\%) on the LIBERO-Object task suite, indicating strong generalization to unseen objects in simulation environments. We conjecture that the generalization capabilities of the CLIP model to novel visual categories~\citep{radford2021learning,fang2023data} are successfully transferred to the LIBERO-Object task suite.

We further analyze the inference efficiency of CLIP-RT+. We use two evaluation metrics: (1) throughput (the number of actions predicted per second) and (2) latency (time to predict an action chunk or single action). By following the setup from \citet{kim2025fine}, we measure the throughput and latency on an NVIDIA A100 GPU. As shown in Table~\ref{tab:libero}, CLIP-RT+ achieves 39$\times$ improved throughput (4.2Hz$\rightarrow$163.8Hz) compared with OpenVLA based on its lightweight design and the action chunking technique. When compared to OpenVLA-OFT using the same action chunk size of 8, CLIP-RT+ improves both throughput and latency by approximately 49\%.

While LIBERO demonstrations are not compatible with language-based action representations due to their low-level, continuous action space nature, we adapt CLIP-RT by adding a simple action prediction module with an L1 regression objective for continuous action representations. This modification enables us to evaluate the core architectural strengths of CLIP-RT—language-based policy pretraining and lightweight design—on a widely used simulation benchmark (LIBERO). The results demonstrate that CLIP-RT remains effective and generalizable, even when applied beyond the scope of language supervision-based robot learning settings.

%% file: sections/06_limitation.tex


\section{Discussion}

\subsection{Summary}

This paper investigates: (1) how non-experts collect robotic data using natural language supervision and (2) how pre-trained vision-language models learn visuomotor policies directly from this supervision. We present CLIP-RT, a new vision-language-action (VLA) model that learns generalizable and transferable policies from natural language supervision. Furthermore, we propose a data collection framework consisting of language-based teleoperation and stochastic trajectory augmentation. Experiments show that CLIP-RT outperforms the state-of-the-art model, OpenVLA by 24\%, in acquiring novel manipulation skills, while using 7x fewer parameters. Furthermore, CLIP-RT can collaborate with humans and large pretrained models by using natural language as an interface, improving generalization and decision-making. Finally, we validate the effectiveness of CLIP-RT in simulated environments with offline, human-teleoperated robot data. We believe that our work represents a promising step towards making robot learning more accessible and scalable, enabling non-experts to teach robots directly in their environments.

\subsection{Limitations and Future Work}


\noindent\textbf{Inherent Limitations in Human Language Supervision.}
Human can provide instructions at varying levels of abstraction—from high-level commands like ``\emph{Pick up the cup}'' to low-level directives such as ``\emph{Rotate the second joint by 10 degrees}''.
Our approach currently assumes that users can offer supervision at an appropriate intermediate level (e.g., \emph{move arm to the right}). 
This assumption may not hold in real-world scenarios, as non-experts might struggle to calibrate the specificity of their instructions.
Addressing this limitation may involve developing adaptive models capable of interpreting instructions across different levels of abstraction or designing a two-stage pipeline that first translates high-level instructions into intermediate commands and subsequently into low-level actions, as demonstrated in \cite{song2023llm,shin2024socratic}.

\noindent\textbf{Lack of Temporal Context.}
Current vision-language-action models, including CLIP-RT, do not predict sequences of actions or consider the history of actions taken. 
This absence of temporal context limits the models' ability to perform tasks that require an understanding of previous actions or states. 
For instance, in a task like \emph{Shake the water bottle}, the robot needs to know whether it has already shaken the bottle or how it should continue shaking. 
Without incorporating action history into the context, the model cannot make informed decisions based on past actions. 
Future research could explore integrating mechanisms that account for temporal sequences, enabling the model to maintain a memory of prior actions and states, such as hierarchical history encoding~\cite{chen2021history}.

\noindent\textbf{Handling Complex Tasks and Long-Term Planning.} The robotic tasks addressed in this paper are relatively short-horizon compared with the complexity and duration of everyday tasks, such as folding laundry~\citep{black2024pi_0}. While CLIP-RT successfully demonstrates diverse manipulation skills — such as opening the trash can and closing the laptop — extending these capabilities to long-horizon tasks requires novel approaches that can handle increased task complexity. One promising strategy for long-horizon task execution involves developing a high-level task planner~\citep{huang2022language,song2023llm,shin2024socratic} that decomposes complex tasks into sequences of primitive skills. For example, a task planner could break down ``set the dinner table'' into subtasks like ``retrieve plates,'' ``place utensils,'' and ``arrange napkins.'' Integrating such planners with CLIP-RT's manipulation skills could execute structured, multi-step tasks.

%% file: sections/07_ack.tex
\section*{Acknowledgements}

We would like to thank Eungseo Kim for his help with robot evaluation in simulated environments.
This work was partly supported by the IITP (RS-2021-II212068-AIHub/10\%, RS-2021-II211343-GSAI/10\%, RS-2022-II220951-LBA/10\%, RS-2022-II220953-PICA/15\%), NRF (RS-2024-00353991-SPARC/15\%, RS-2023-00274280-HEI/10\%), KEIT (RS-2024-00423940/10\%), KRIT (KRIT-CT-23-003/10\%), and Gwangju Metropolitan City (Artificial intelligence industrial convergence cluster development project/10\%) grant funded by the Korean government.

%% file: sections/08_appendix.tex
\begin{appendices}
\onecolumn
\input{sections/appx/01_nls}

\input{sections/appx/02_lookup_table}

\input{sections/appx/03_llm_prompt}
\twocolumn
\input{sections/appx/04_robot_platform}

\input{sections/appx/06_std}

\input{sections/appx/07_supp_exp}

\input{sections/appx/08_task_details}

\input{sections/appx/11_embedding_vis}
\input{sections/appx/09_ensemble_additional}
\twocolumn
\end{appendices}


%% file: sections/appx/01_nls.tex
\section{Natural Language Supervisions}
\label{tab:nls}
We define 50 types of natural language supervision as follows. These include shifting the end-effector position, orientation, and gripper state. For position control, we categorize the supervision into four granularities (1cm, 5cm, 10cm, and 20cm). Likewise, orientation adjustments are categorized into four granularities (5 degrees, 15 degrees, 45 degrees, and 90 degrees). We leverage GPT-4~\cite{gpt4} to augment these 50 types of supervisions into 899 natural language supervisions. For pretraining, we transform low-level end-effector actions in the Open X-Embodiment dataset~\cite{padalkar2023open} into natural language supervision. We identify the dominant axis and corresponding value of each end-effector action and directly map this information to one of the following types of natural language supervisions. For example, an end-effector action that moves the robot arm backward by 6.5cm, this action is mapped to the third type, ``move arm back by 5cm.'' The final natural language supervision is selected from 899 generated instructions (\textit{e.g.,} ``pull arm back 5cm'').

\begin{tcolorbox}[
    enhanced,
    breakable, 
    colback=black!5!white, 
    colframe=black!50!black]
\begin{multicols}{2}
\begin{enumerate}
[
    topsep=-0.2pt, 
    partopsep=-0.2pt, 
    itemsep=-0.2pt, 
    parsep=-0.2pt, 
    leftmargin=20pt
]
\item move arm back by 20cm
\item move arm back by 10cm
\item move arm back by 5cm
\item move arm back by 1cm
\item move arm forward by 1cm
\item move arm forward by 5cm
\item move arm forward by 10cm
\item move arm forward by 20cm
\item move arm to the right by 20cm
\item move arm to the right by 10cm
\item move arm to the right by 5cm
\item move arm to the right by 1cm
\item move arm to the left by 1cm
\item move arm to the left by 5cm
\item move arm to the left by 10cm
\item move arm to the left by 20cm
\item lower arm by 20cm
\item lower arm by 10cm
\item lower arm by 5cm
\item lower arm by 1cm
\item raise arm up by 1cm
\item raise arm up by 5cm
\item raise arm up by 10cm
\item raise arm up by 20cm
\item roll arm 90 degrees counterclockwise
\item roll arm 45 degrees counterclockwise
\item roll arm 15 degrees counterclockwise
\item roll arm 5 degrees counterclockwise
\item roll arm 5 degrees clockwise
\item roll arm 15 degrees clockwise
\item roll arm 45 degrees clockwise
\item roll arm 90 degrees clockwise
\item tilt arm up 90 degrees
\item tilt arm up 45 degrees
\item tilt arm up 15 degrees
\item tilt arm up 5 degrees
\item tilt arm down 5 degrees
\item tilt arm down 15 degrees
\item tilt arm down 45 degrees
\item tilt arm down 90 degrees
\item yaw arm 90 degrees counterclockwise
\item yaw arm 45 degrees counterclockwise
\item yaw arm 15 degrees counterclockwise
\item yaw arm 5 degrees counterclockwise
\item yaw arm 5 degrees clockwise
\item yaw arm 15 degrees clockwise
\item yaw arm 45 degrees clockwise
\item yaw arm 90 degrees clockwise
\item close the gripper
\item open the gripper
\end{enumerate}
\end{multicols}
\end{tcolorbox}

%% file: sections/appx/02_lookup_table.tex
\section{A Lookup Table for Closed-Loop Robot Control}
\label{tab:lookup}
We use a lookup table for closed-loop robot control as follows. Based on 50 natural language supervisions above, we add 8 natural language supervisions regarding the gripper rotation, resulting in 58 natural language supervisions. This is because it is often more intuitive to guide robots to rotate the gripper rather than rolling or yawing the arm. Accordingly, we add an additional dimension to the original 7D end-effector actions to represent the gripper rotation, resulting in 8D end-effector actions. Adding one dimension to the end-effector actions does not affect CLIP-RT since our model learns robotic policies directly from natural language supervision. However, this affects existing models like OpenVLA~\cite{kim24openvla} which are pre-trained to predict 7-dimensional actions. To make it compatible, we transform 8D commands to 7D commands when training the OpenVLA model by expressing the gripper rotation as rolling or yawing the robot arm.

\begin{tcolorbox}[breakable, colback=black!5!white, colframe=black!50!black]
\{
\begin{enumerate}[label=, topsep=-0.2pt, partopsep=-0.2pt, itemsep=-0.2pt, parsep=-0.2pt, leftmargin=20pt]
\item move arm back by 20cm: [-0.2, 0.0, 0.0, 0.0, 0.0, 0.0, 0.0, -1.0],
\item move arm back by 10cm: [-0.1, 0.0, 0.0, 0.0, 0.0, 0.0, 0.0, -1.0],
\item move arm back by 5cm: [-0.05, 0.0, 0.0, 0.0, 0.0, 0.0, 0.0, -1.0],
\item move arm back by 1cm: [-0.01, 0.0, 0.0, 0.0, 0.0, 0.0, 0.0, -1.0],
\item move arm forward by 1cm: [0.01, 0.0, 0.0, 0.0, 0.0, 0.0, 0.0, -1.0],
\item move arm forward by 5cm: [0.05, 0.0, 0.0, 0.0, 0.0, 0.0, 0.0, -1.0],
\item move arm forward by 10cm: [0.1, 0.0, 0.0, 0.0, 0.0, 0.0, 0.0, -1.0],
\item move arm forward by 20cm: [0.2, 0.0, 0.0, 0.0, 0.0, 0.0, 0.0, -1.0],
\item move arm to the right by 20cm: [0.0, -0.2, 0.0, 0.0, 0.0, 0.0, 0.0, -1.0],
\item move arm to the right by 10cm: [0.0, -0.1, 0.0, 0.0, 0.0, 0.0, 0.0, -1.0],
\item move arm to the right by 5cm: [0.0, -0.05, 0.0, 0.0, 0.0, 0.0, 0.0, -1.0],
\item move arm to the right by 1cm: [0.0, -0.01, 0.0, 0.0, 0.0, 0.0, 0.0, -1.0],
\item move arm to the left by 1cm: [0.0, 0.01, 0.0, 0.0, 0.0, 0.0, 0.0, -1.0],
\item move arm to the left by 5cm: [0.0, 0.05, 0.0, 0.0, 0.0, 0.0, 0.0, -1.0],
\item move arm to the left by 10cm: [0.0, 0.1, 0.0, 0.0, 0.0, 0.0, 0.0, -1.0],
\item move arm to the left by 20cm: [0.0, 0.2, 0.0, 0.0, 0.0, 0.0, 0.0, -1.0],
\item lower arm by 20cm: [0.0, 0.0, -0.2, 0.0, 0.0, 0.0, 0.0, -1.0],
\item lower arm by 10cm: [0.0, 0.0, -0.1, 0.0, 0.0, 0.0, 0.0, -1.0],
\item lower arm by 5cm: [0.0, 0.0, -0.05, 0.0, 0.0, 0.0, 0.0, -1.0],
\item lower arm by 1cm: [0.0, 0.0, -0.01, 0.0, 0.0, 0.0, 0.0, -1.0],
\item raise arm up by 1cm: [0.0, 0.0, 0.01, 0.0, 0.0, 0.0, 0.0, -1.0],
\item raise arm up by 5cm: [0.0, 0.0, 0.05, 0.0, 0.0, 0.0, 0.0, -1.0],
\item raise arm up by 10cm: [0.0, 0.0, 0.1, 0.0, 0.0, 0.0, 0.0, -1.0],
\item raise arm up by 20cm: [0.0, 0.0, 0.2, 0.0, 0.0, 0.0, 0.0, -1.0],
\item roll arm 90 degrees counterclockwise: [0.0, 0.0, 0.0, -1.5708, 0.0, 0.0, 0.0, -1.0],
\item roll arm 45 degrees counterclockwise: [0.0, 0.0, 0.0, -0.7854, 0.0, 0.0, 0.0, -1.0],
\item roll arm 15 degrees counterclockwise: [0.0, 0.0, 0.0, -0.2618, 0.0, 0.0, 0.0, -1.0],
\item roll arm 5 degrees counterclockwise: [0.0, 0.0, 0.0, -0.0872, 0.0, 0.0, 0.0, -1.0],
\item roll arm 5 degrees clockwise: [0.0, 0.0, 0.0, 0.0872, 0.0, 0.0, 0.0, -1.0],
\item roll arm 15 degrees clockwise: [0.0, 0.0, 0.0, 0.2618, 0.0, 0.0, 0.0, -1.0],
\item roll arm 45 degrees clockwise: [0.0, 0.0, 0.0, 0.7854, 0.0, 0.0, 0.0, -1.0],
\item roll arm 90 degrees clockwise: [0.0, 0.0, 0.0, 1.5708, 0.0, 0.0, 0.0, -1.0],
\item tilt arm up 90 degrees: [0.0, 0.0, 0.0, 0.0, -1.5708, 0.0, 0.0, -1.0],
\item tilt arm up 45 degrees: [0.0, 0.0, 0.0, 0.0, -0.7854, 0.0, 0.0, -1.0],
\item tilt arm up 15 degrees: [0.0, 0.0, 0.0, 0.0, -0.2618, 0.0, 0.0, -1.0],
\item tilt arm up 5 degrees: [0.0, 0.0, 0.0, 0.0, -0.0872, 0.0, 0.0, -1.0],
\item tilt arm down 5 degrees: [0.0, 0.0, 0.0, 0.0, 0.0872, 0.0, 0.0, -1.0],
\item tilt arm down 15 degrees: [0.0, 0.0, 0.0, 0.0, 0.2618, 0.0, 0.0, -1.0],
\item tilt arm down 45 degrees: [0.0, 0.0, 0.0, 0.0, 0.7854, 0.0, 0.0, -1.0],
\item tilt arm down 90 degrees: [0.0, 0.0, 0.0, 0.0, 1.5708, 0.0, 0.0, -1.0],
\item yaw arm 90 degrees counterclockwise: [0.0, 0.0, 0.0, 0.0, 0.0, -1.5708, 0.0, -1.0],
\item yaw arm 45 degrees counterclockwise: [0.0, 0.0, 0.0, 0.0, 0.0, -0.7854, 0.0, -1.0],
\item yaw arm 15 degrees counterclockwise: [0.0, 0.0, 0.0, 0.0, 0.0, -0.2618, 0.0, -1.0],
\item yaw arm 5 degrees counterclockwise: [0.0, 0.0, 0.0, 0.0, 0.0, -0.0872, 0.0, -1.0],
\item yaw arm 5 degrees clockwise: [0.0, 0.0, 0.0, 0.0, 0.0, 0.0872, 0.0, -1.0],
\item yaw arm 15 degrees clockwise: [0.0, 0.0, 0.0, 0.0, 0.0, 0.2618, 0.0, -1.0],
\item yaw arm 45 degrees clockwise: [0.0, 0.0, 0.0, 0.0, 0.0, 0.7854, 0.0, -1.0],
\item yaw arm 90 degrees clockwise: [0.0, 0.0, 0.0, 0.0, 0.0, 1.5708, 0.0, -1.0],
\item rotate gripper 90 degrees counterclockwise: [0.0, 0.0, 0.0, 0.0, 0.0, 0.0, -1.5708, -1.0],
\item rotate gripper 45 degrees counterclockwise: [0.0, 0.0, 0.0, 0.0, 0.0, 0.0, -0.7854, -1.0],
\item rotate gripper 15 degrees counterclockwise: [0.0, 0.0, 0.0, 0.0, 0.0, 0.0, -0.2618, -1.0],
\item rotate gripper 5 degrees counterclockwise: [0.0, 0.0, 0.0, 0.0, 0.0, 0.0, -0.0872, -1.0],
\item rotate gripper 5 degrees clockwise: [0.0, 0.0, 0.0, 0.0, 0.0, 0.0, 0.0872, -1.0],
\item rotate gripper 15 degrees clockwise: [0.0, 0.0, 0.0, 0.0, 0.0, 0.0, 0.2618, -1.0],
\item rotate gripper 45 degrees clockwise: [0.0, 0.0, 0.0, 0.0, 0.0, 0.0, 0.7854, -1.0],
\item rotate gripper 90 degrees clockwise: [0.0, 0.0, 0.0, 0.0, 0.0, 0.0, 1.5708, -1.0],
\item close the gripper: [0.0, 0.0, 0.0, 0.0, 0.0, 0.0, 0.0, 0.0]
\item open the gripper: [0.0, 0.0, 0.0, 0.0, 0.0, 0.0, 0.0, 1.0],
\end{enumerate}
\}
\end{tcolorbox}
\clearpage

%% file: sections/appx/03_llm_prompt.tex
\section{LLM Prompt for Language-Based Teleoperation}
\label{tab:prompt}
\begin{tcolorbox}[colback=black!5!white, colframe=black!50!black]

You are a generalist agent who can control a physical robot arm with a two-finger gripper, given natural language supervision from humans. Please return the desired output by referring to the following explanation.\vspace{0.3cm}

\textbf{TASK DESCRIPTION: }

The task you should perform is to translate the natural language supervision from humans into the corresponding robot end effector command. Language supervision entails diverse actions: (1) displacing the end effector's 3D Cartesian coordinates or poses and (2) directly rotating the gripper. \vspace{0.3cm}
    
\textbf{OUTPUT FORMAT:}

You should output the end effector command, which is a list of length seven. The first six elements correspond to the standard end effector commands. Specifically, the first three elements of the command represent delta Cartesian coordinates of the end effector (i.e., modified x, y, and z coordinates), and the next three correspond to the orientation of the end effector (i.e., modified roll, pitch, and yaw). In addition to the list of length six, we define the last element of the end effector command as the robotic arm's last joint angles to directly rotate gripper. Please note that you should output the list of length seven without detailed explanation.\vspace{0.3cm}
    
\textbf{ENVIRONMENT SETUP:}

The physical robot arm is standing on the table, and the gripper is mounted at the end of the robotic arm. The 3D Cartesian coordinate system of the environment is as follows:
\begin{enumerate}[topsep=0pt, partopsep=0pt, itemsep=0pt, parsep=0pt]
\item The x-axis is in the depth direction, increasing away from you.
\item The y-axis is in the horizontal direction, increasing to the left.
\item The z-axis is in the vertical direction, increasing upwards.
\end{enumerate}
\vspace{0.3cm}

\textbf{RULES:}

Please note that the following rules when predicting the end effector command:
\begin{enumerate}[topsep=0pt, partopsep=0pt, itemsep=0pt, parsep=0pt]
\item The units for the Cartesian coordinate system are meters.
\item The units for the roll, pitch, and yaw are degrees, from -90 to 90 degrees.
\item The joint angles of the gripper also ranges from -90 to 90 degrees.
\item Positive rotation values represent clockwise rotation, and negative rotation values represent counterclockwise rotation.
\item The end effector gripper has two fingers, and the fingers are opened and pointing downward in the initial state (i.e., parallel to the z-axis). You should predict the delta roll, pitch, and yaw based on the initial orientation.
\item If the natural language supervision does not seem relevant to the end effector commands, you should output a list of zero values.
\end{enumerate}
\vspace{0.3cm}

\textbf{EXAMPLE:}

Here are a few examples for natural language supervision and the end effector command:

\begin{enumerate}[topsep=0pt, partopsep=0pt, itemsep=0pt, parsep=0pt]
\item move to the right: [0.0, -0.1, 0.0, 0.0, 0.0, 0.0, 0.0]
\item move forward a bit: [0.05, 0.0, 0.0, 0.0, 0.0, 0.0, 0.0]
\item lower arm a tiny bit: [0.0, 0.0, -0.01, 0.0, 0.0, 0.0, 0.0]
\item raise arm up a lot: [0.0, 0.0, 0.2, 0.0, 0.0, 0.0, 0.0]
\item roll arm to the left a bit: [0.0, 0.0, 0.0, 15.0, 0.0, 0.0, 0.0]
\item tilt end effector up a lot: [0.0, 0.0, 0.0, 0.0, -90.0, 0.0, 0.0]
\item yaw arm to the left a tiny bit: [0.0, 0.0, 0.0, 0.0, 0.0, -5.0, 0.0]
\item rotate gripper 45 degrees clockwise: [0.0, 0.0, 0.0, 0.0, 0.0, 0.0, 45.0]
\item close the gripper : [0.0, 0.0, 0.0, 0.0, 0.0, 0.0, 0.0].
\end{enumerate}
\vspace{0.3cm} 

Based on the description above, please infer the end effector command for the natural language supervision, $\left\{\text{supervision}\right\}$.
\end{tcolorbox}
\clearpage

%% file: sections/appx/04_robot_platform.tex




\section{Robot Platform}
\label{sec:robot_platform}
\begin{figure}[H]
    \centering
    \includegraphics[width=0.68\columnwidth]{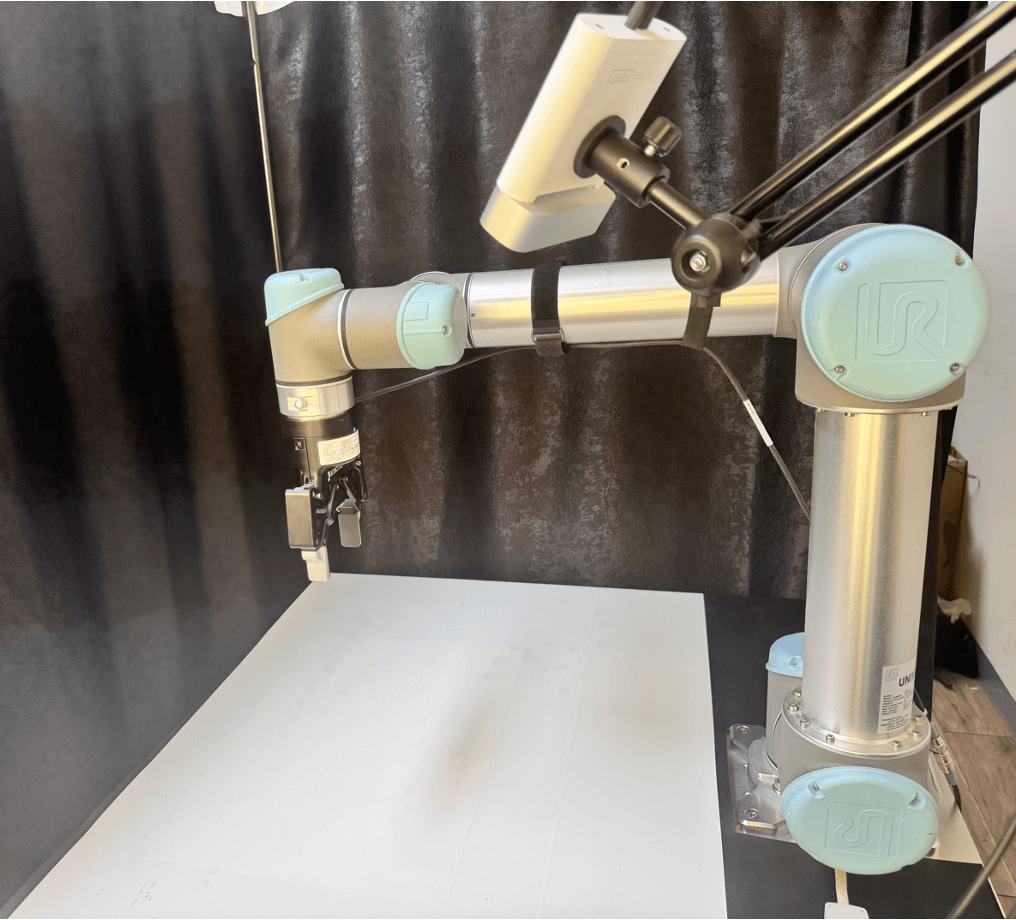}
    \caption{
    \textbf{Robot Platform} 
    }
    \label{fig:exp_setup}
\end{figure}

We constructed a tabletop environment to perform manipulation tasks, as shown in Figure~\ref{fig:exp_setup}. The experiments are carried out using a 6-DoF Universal Robots UR5 physical robotic arm equipped with a two-fingered gripper. All episodes begin from a standard home pose, as shown in the figure, and objects are placed within the white area to ensure that they are within the robot’s reachable workspace. For visual input, we utilized an Azure Kinect DK, which provides an RGB image of the scene. The camera position remains fixed throughout all experiments, positioned to the left and slightly behind the robot arm to ensure consistent visual perspectives across tasks.

%% file: sections/appx/06_std.tex
\section{Algorithm of Stochastic Trajectory Augmentation (STA)}
\label{sec:appendix_std_algorithm}

Algorithm~\ref{alg:std} presents the \textit{Stochastic Trajectory Augmentation (STA)} to augment an expert trajectory $\mathcal{T}_{\text{expert}}$. The goal of this algorithm is to enhance the robot's generalization and resilience by introducing variability in the actions between waypoints and training the robot to recover from deviations. The algorithm inputs $\mathcal{T}_{\text{expert}}$ and outputs an augmented trajectory $\mathcal{T}_{\text{aug}}$.

In the Diversification phase, for each segment between consecutive waypoints $w_k$ and $w_{k+1}$, the algorithm computes the cumulative action $\Delta \mathbf{a}$ required to move from $w_k$ to $w_{k+1}$:

\begin{equation} \Delta \mathbf{a} = \sum_{t = t_k}^{t_{k+1}} \mathbf{a}_t \end{equation}

where $\mathbf{a}_t$ are the actions at each time step $t$ between the waypoints $w_k$ and $w_{k+1}$, and $t_k$ and $t_{k+1}$ are the corresponding time steps. This cumulative action includes position changes $(\Delta x, \Delta y, \Delta z)$ and rotations $(\Delta r, \Delta p, \Delta y, \Delta g)$, representing roll, pitch, yaw, and gripper rotation. The remaining positional action $\Delta \mathbf{a}_{\text{rem}}$ is initialized with $(\Delta x, \Delta y, \Delta z)$.
The algorithm then iteratively samples feasible action increments $\delta \mathbf{a}_n$ from a set of values that do not exceed the remaining action and have the same direction (positive or negative), selected from a predefined set $S = \{0.01, 0.05, 0.1\}$ (representing 1, 5, and 10 cm). STA generates diversified trajectories by iteratively sampling action increments $\delta \mathbf{a}_n$ such that:

\begin{equation} \Delta \mathbf{a} = \sum_{n} \delta \mathbf{a}_n \end{equation}

ensuring that the new intermediate actions sum up to the cumulative action $\Delta \mathbf{a}$. These increments are appended to the augmented trajectory $\mathcal{T}_{\text{aug}}$, gradually reducing the remaining action until it is below a small threshold $\epsilon$. This process introduces variability in the intermediate actions while maintaining the overall trajectory, thereby improving the robot's robustness and ability to generalize to unseen scenarios.

In the Recovery phase, when the gripper is close to the waypoint, the algorithm introduces deliberate deviations to train the robot's recovery capability. Creates a deviation action $\delta \mathbf{a}_{\text{dev}}$ by sampling $S$ in the dimensions $x$ and $y$ and setting the $z$ component to the absolute value of the remaining action $\Delta \mathbf{a}_{\text{rem},z}$. Since the gripper is often close to the table, the $z$ component is made positive for safety. 
The sampling continues until a deviation action moves the gripper farther away from the waypoint. The recovery action $\delta \mathbf{a}_{\text{rec}}$ is then calculated as the negation of the deviation action:

\begin{equation} 
\delta \mathbf{a}_{\text{dev}} = [\delta a_{\text{dev}, x}, \delta a_{\text{dev}, y}, \delta a_{\text{dev}, z}, 0, 0, 0, 0]\end{equation}
\begin{equation}
    \quad \delta \mathbf{a}_{\text{rec}} = -\delta \mathbf{a}_{\text{dev}}
\end{equation}
Both the deviation and recovery actions are appended to $\mathcal{T}_{\text{aug}}$. However, note that the deviation action is omitted in the training data. 
Training with the recovery action enhances the robot's resilience, enabling it to handle unexpected disturbances and recover efficiently during task execution.

Finally, if there are rotational changes $(\Delta r, \Delta p, \Delta y, \Delta g)$ in the segment, the algorithm appends these actions to the augmented trajectory. By combining the Diversification phase and the Recovery phase, the algorithm generates a diversified set of trajectories that not only cover various valid action sequences leading to the goal but also prepare the robot to handle unexpected disturbances. This comprehensive augmentation enhances the robot's robustness and ability to generalize to new scenarios.

The robot then executes the augmented trajectory $\mathcal{T}_{\text{aug}}$ and collects images for model training. Actions from the STA phase, $\delta \mathbf{a}$, and the recovery actions of the Recovery phase, $\delta \mathbf{a}_{\text{rec}}$, are used for model training, while the deviation actions, $\delta \mathbf{a}_{\text{dev}}$, are omitted.

\onecolumn

\begin{algorithm}[t] 
\caption{Algorithm of Stochastic Trajectory Augmentation (STA)} 
\label{alg:std}
\begin{algorithmic}[1] 
\Require Expert trajectory $\mathcal{T}_{\text{expert}}$ divided into segments between waypoints $\{w_1, w_2, \ldots, w_n\}$ 
\Ensure Augmented trajectory $\mathcal{T}_{\text{aug}}$ with diversified actions and recovery actions
\State Define sample sizes $S = \{0.01, 0.05, 0.1\}$

\For{each segment between consecutive waypoints $w_k, w_{k+1}$} 
\State Compute cumulative action $\Delta \mathbf{a} = \sum_{t=t_k}^{t_{k+1}} \mathbf{a}_t = [\Delta x, \Delta y, \Delta z, \Delta r, \Delta p, \Delta y, \Delta g]$
\State Initialize remaining action $\Delta \mathbf{a}_{\text{rem}} \gets [\Delta x, \Delta y, \Delta z]$ 
\State \textbf{Diversification Phase}
\While{$\|\Delta \mathbf{a}_{\text{rem}}\| > \epsilon$} 
    \State Initialize sampled action $\delta \mathbf{a} \gets [0, 0, 0, 0, 0, 0, 0]$
    \For{each position dimension $i \in \{x, y, z\}$}
        \If{$\Delta a_{\text{rem}, i} < 0$}
            \State $V_i \gets \{ -s \mid s \in S, s \leq |\Delta a_{\text{rem}, i}| \}$
        \Else
            \State $V_i \gets \{ s \mid s \in S, s \leq \Delta a_{\text{rem}, i} \}$
        \EndIf
        \State $v_i \sim \text{Uniform}(V_i)$ (or set $v_i = 0$ if $V_i$ is empty)
        \State Update $\delta a_i \gets v_i$
    \EndFor
    \State Append $\delta \mathbf{a}$ to $\mathcal{T}_{\text{aug}}$
    \State Update $\Delta \mathbf{a}_{\text{rem}} \gets \Delta \mathbf{a}_{\text{rem}} - [\delta a_x, \delta a_y, \delta a_z]$

\State \textbf{Recovery Phase}
\If{$\|\Delta \mathbf{a}_{\text{rem}}\| < \text{threshold}$}
    \State Create Deviation Action $\delta \mathbf{a}_{\text{dev}} \gets [0, 0, 0, 0, 0, 0, 0]$
    \While{$\|\Delta \mathbf{a}_{\text{rem}}\| < \|\Delta \mathbf{a}_{\text{rem}} + \delta \mathbf{a}_{\text{dev}}\|$}
    \For{each position dimension $i \in \{x, y\}$}
        \If{$\Delta a_{\text{rem}, i} < 0$}
            \State $\delta \mathbf{a}_{\text{dev},i} \sim \text{Uniform}(\{ -s \mid s \in S \})$
        \Else
            \State $\delta \mathbf{a}_{\text{dev},i} \sim \text{Uniform}(S)$
        \EndIf
    \EndFor
    \State $\delta \mathbf{a}_{\text{dev},z} \gets \mid \Delta \mathbf{a}_{\text{rem},z} \mid$
    \EndWhile
    \State Recovery Action $\delta \mathbf{a}_{\text{rec}} \gets -\delta \mathbf{a}_{\text{dev}}$
    \State Append $\delta \mathbf{a}_{\text{dev}}$ and $\delta \mathbf{a}_{\text{rec}}$ to $\mathcal{T}_{\text{aug}}$
\EndIf
\EndWhile
\If{$\Delta r, \Delta p, \Delta y, \Delta g$ are non-zero}
    \State Append rotation action $\delta \mathbf{a}_{\text{rot}} = [0, 0, 0, \Delta r, \Delta p, \Delta y, \Delta g]$ to $\mathcal{T}_{\text{aug}}$
\EndIf
\EndFor
\State \textbf{Output:} Augmented trajectory $\mathcal{T}_{\text{aug}}$
\end{algorithmic} 
\end{algorithm}

\twocolumn

%% file: sections/appx/07_supp_exp.tex
\onecolumn
\section{Supplementary Experiments}

\subsection{Additional Analysis Comparing CLIP-RT with OpenVLA}
\label{ssec:additional_analysis}
In this section, we highlight several interesting phenomena observed when contrasting CLIP-RT and OpenVLA.

\begin{figure}[!t]
\centering
\includegraphics[width=0.4\columnwidth]{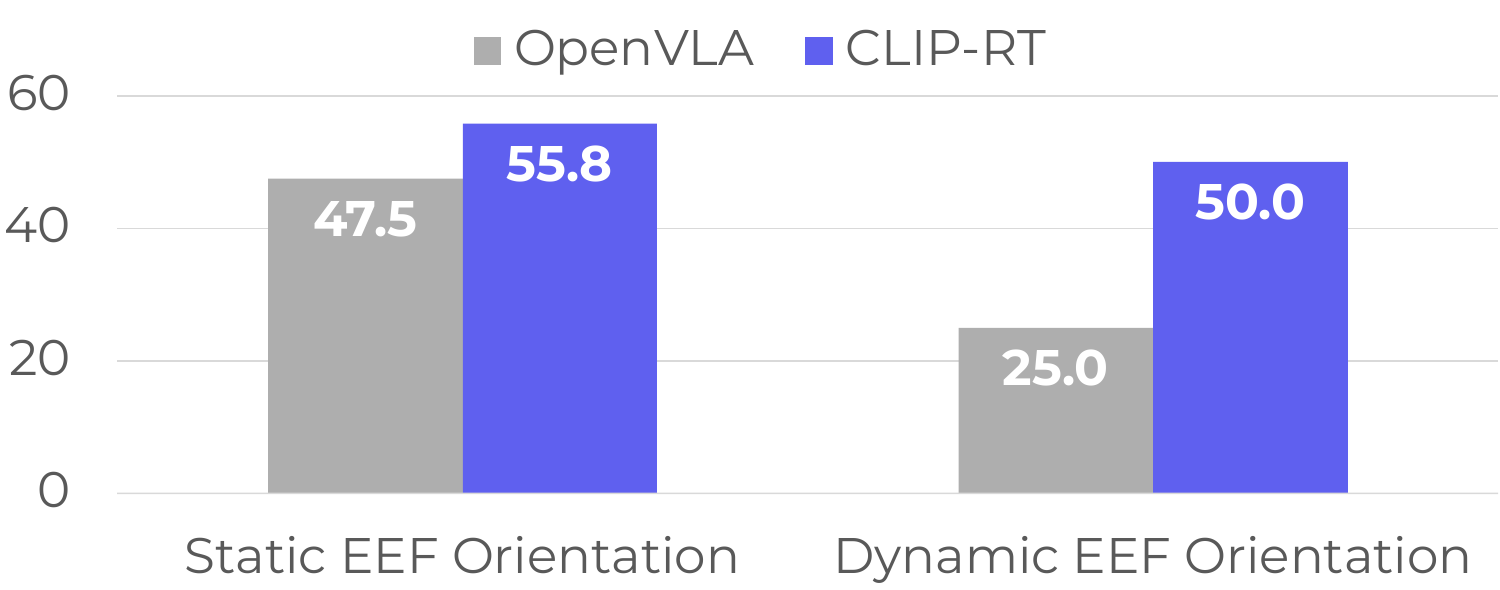}
\caption{\textbf{Comparison of Static \& Dynamic End-Effector Orientation.}}
\label{fig:eef_comparison}
\end{figure}

Figure~\ref{fig:eef_comparison} categorizes the tasks by the degree of end-effector orientation required. In \emph{Static EEF Orientation tasks}, the robot maintains a stable orientation, while in \emph{Dynamic EEF Orientation tasks} roll----pitch----yaw adjustments are demanded. Although CLIP-RT achieves a success rate of 8\% higher than OpenVLA in Static tasks, that margin increases to 25\% when orientation changes are required. These results demonstrate that complex orientation control amplifies the advantages of CLIP-RT, highlighting its robust performance even in demanding manipulation scenarios.

A detailed comparison of individual \emph{Common} tasks further distinguishes the performance of CLIP-RT and OpenVLA (see Figure~\ref{fig:result_performance}. For relatively simpler tasks such as \emph{Point}, \emph{Pull}, \emph{Place}, \emph{Pick}, and \emph{Push}, both models perform comparably, with OpenVLA slightly outperforming CLIP-RT. This suggests that for tasks requiring straightforward action mappings, OpenVLA is highly effective. However, in more challenging tasks such as \emph{Flip}, \emph{Knock Over}, and \emph{Slide}, CLIP-RT significantly outperforms OpenVLA. These tasks involve a higher level of reasoning, such as determining whether an object is upside down (\emph{Flip}) or whether the cup is sufficiently tilted (\emph{Knock Over}).
We conjecture that CLIP-RT’s discriminative approach, which selects actions by directly matching the natural language representation of the desired action to the current context, confers an advantage in these complex tasks. Using the rich semantics available in language and vision, CLIP-RT can make more nuanced decisions. Conversely, OpenVLA’s generative approach to action prediction may be more prone to inaccuracies in these scenarios, as producing precise actions becomes increasingly challenging with higher task complexity.

\begin{figure*}[!t]
    \centering
    \includegraphics[width=1\textwidth]{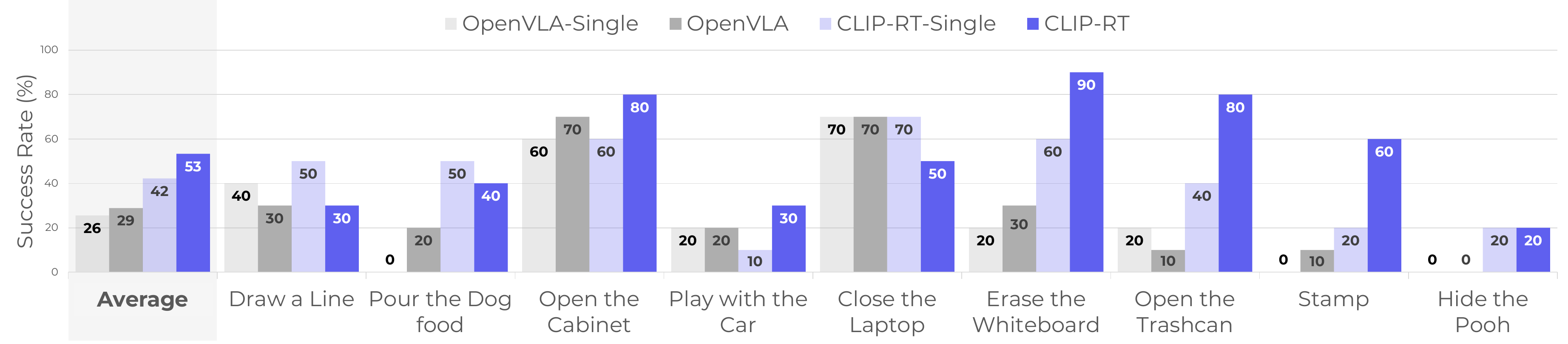}
    \caption{
    \textbf{Multi-Task and Single-Task training on the Novel tasks}
    }
    \label{fig:novel_single}
\end{figure*}

\begin{figure*}[!t]
    \centering
    \includegraphics[width=1\textwidth]{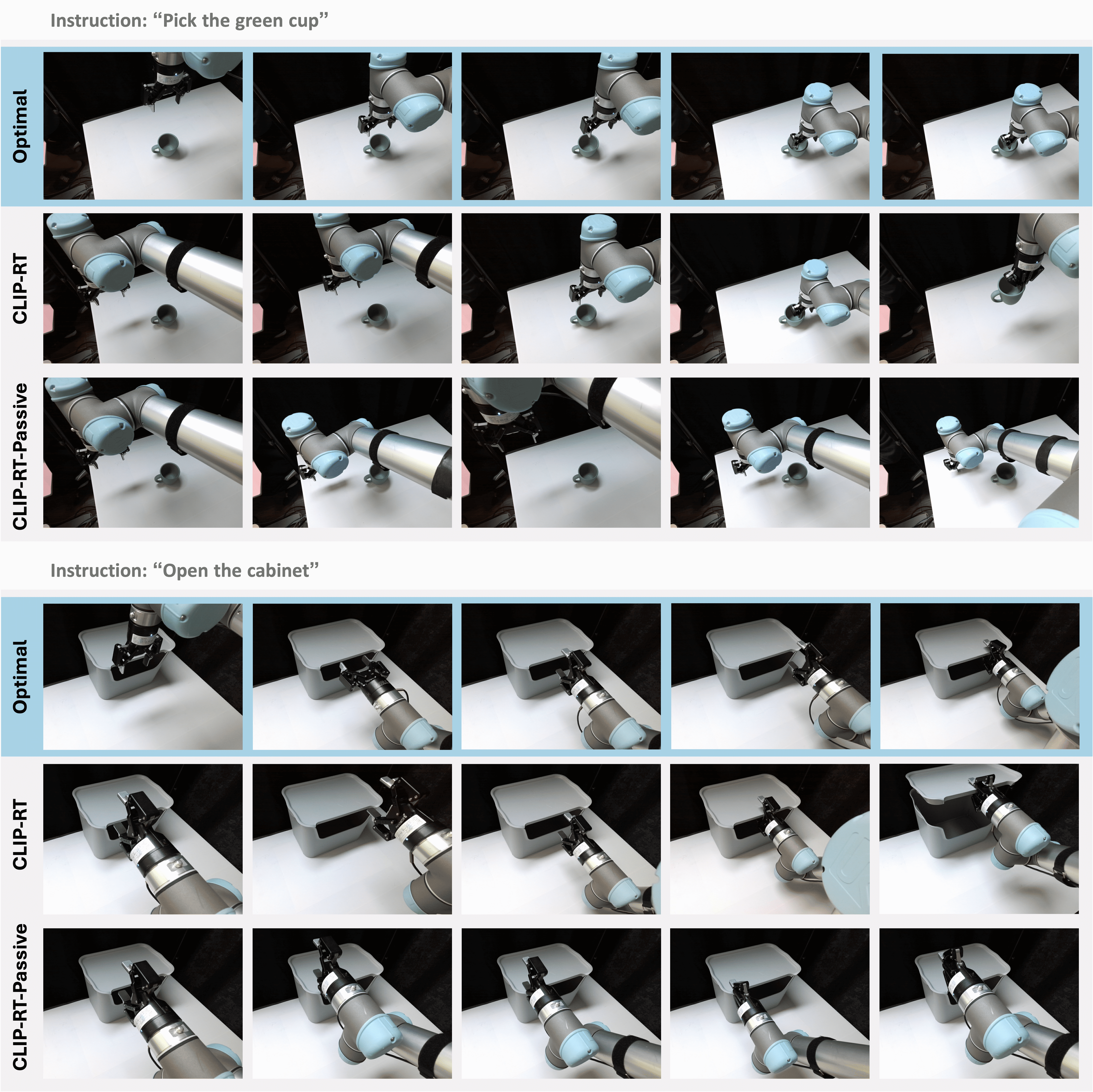}
    \caption{
    \textbf{Qualitative Evaluation of Trajectory Resilience.}    
    this figure illustrates the execution of two tasks: “Pick the green cup” and “Open the cabinet”. 
    It presents three trajectories for each task: (1) the optimal trajectory starting from a standardized home pose, representing an expert trajectory collected through language-based teleoperation, and (2) and (3) the trajectories inferred by CLIP-RT and CLIP-RT-Passive, respectively, starting from a non-optimal point deviated from the expert path.
    }
    \label{fig:qualitative_samples}
\end{figure*}

\subsection{Qualitative Evaluation of Trajectory Resilience}
\label{ssec:qualitative_std}

Figure~\ref{fig:qualitative_samples} shows the robot's trajectory using CLIP-RT and CLIP-RT-Passive models, both starting tasks from non-optimal points, deviating from the optimal trajectory. As depicted, CLIP-RT effectively recovers and aligns itself back to the optimal path, successfully completing the tasks. In contrast, CLIP-RT-Passive, which omits the Stochastic Trajectory Augmentation (STA), struggles to execute both instructions, underscoring the critical role of STA in enhancing the model's performance.

The blue box displays the expert trajectory, illustrating that a model trained solely on expert data lacks the capability to respond effectively when starting from nonoptimal positions. By incorporating deviations from optimal trajectories (as shown in Figure~\ref{fig:std}-(d) and (e)), the model gains the ability to recover from deviated states that cannot be visited through language-based teleoperation.
Stochastic Trajectory Augmentation (STA) thus allows the model to learn from diverse trajectories, enhancing its robustness and generalization capabilities, which are essential for adapting to real-world scenarios.



{\subsection{Preliminary Evaluation on Long‑Horizon Skill Composition}
\label{subsec:long_horizon}

\paragraph{Motivation}
Most prior work evaluates policies on \emph{single‑skill} instructions that exactly match the training distribution (e.g., ``open the trashcan’’ or ``pick up the blue block’’).
More importantly, many policies have \emph{no runtime channel} for human guidance, so failure at one sub‑goal often ends the episode.
By contrast, CLIP‑RT allows an operator to step in verbally when needed.
We therefore ask: given this language interface, \textit{how much human intervention does CLIP‑RT actually require to carry out unseen composite commands} such as
``open the trashcan \emph{and} place the blue block in it’’?

\paragraph{Experimental setup}
We form three two‑step tasks by concatenating skills present during training and record both the success rate (SR) and the mean number of human interventions.  
An intervention is logged whenever the operator judges that the robot has strayed from the intended behavior and issues a corrective natural‑language command.  
Each composite task is run for five trials on the same UR5 platform used throughout the paper; the model itself is \emph{frozen}—no further fine‑tuning or prompt engineering.

\begin{table}[h]
\centering
\small
{
\begin{tabular}{lcc}
\toprule
Composite task & SR (\%) & Interventions $\downarrow$\\
\midrule
Open cabinet \& place cup        & 100 & $7.8\!\pm\!2.7$ \\
Open trashcan \& drop block      & 100 & $7.0\!\pm\!1.1$ \\
Draw line \& erase whiteboard    & 100 & $4.2\!\pm\!0.8$ \\
\bottomrule
\end{tabular}
}
\caption{Long‑horizon evaluation}
\label{tab:long_horizon}
\vspace{-0.6em}
\end{table}

\paragraph{Findings}
As Table~\ref{tab:long_horizon} indicates, CLIP‑RT achieves a 100\% success rate on all three composite tasks.  
Even so, each episode still needs about 4–8 brief interventions, mainly at two points: (i) the very start, when the policy must be told which sub‑task to tackle first, and (ii) hand‑off, where a short verbal cue is needed to launch the next sub‑task.

\paragraph{Implications for scalability}
While our findings demonstrate that CLIP‑RT can execute modestly longer, two‑step commands without additional training, truly long‑horizon goals, e.g., ``clean my room'', remain out of reach. 
To scale further, we envision a hierarchical approach in which CLIP‑RT serves as a fast \emph{System‑1} controller, while a complementary \emph{System‑2} planner reasons over task sequences and maintains global coherence~\citep{ahn2022can}.

\subsection{Effect of Action Space Discretization on CLIP-RT}
\label{ssec:libero_ablation}

To investigate the generality of CLIP-RT's action space design, we evaluate its performance on the LIBERO-Spatial benchmark, which provides continuous 7D end-effector actions as training data. Since CLIP-RT is designed to learn from language-supervised data, where each action corresponds to a discrete motion primitive (e.g., ``move arm to the right by 10cm''), it cannot directly consume continuous action trajectories. Therefore, we quantize LIBERO’s continuous action space to align it with CLIP-RT’s discrete, language-aligned format.

To simulate CLIP-RT's original training setup, we discretize the continuous training data by performing \textit{k}-means clustering per axis, replacing each continuous value with its nearest cluster center. We then train CLIP-RT on these discretized versions of the LIBERO dataset using varying levels of discretization granularity: $k=8$, $16$, $32$, $64$, and $128$. This design allows us to examine whether finer-grained action representations translate into improved performance.

{
\begin{figure*}[!b]
    \centering
    \includegraphics[width=0.6\linewidth]{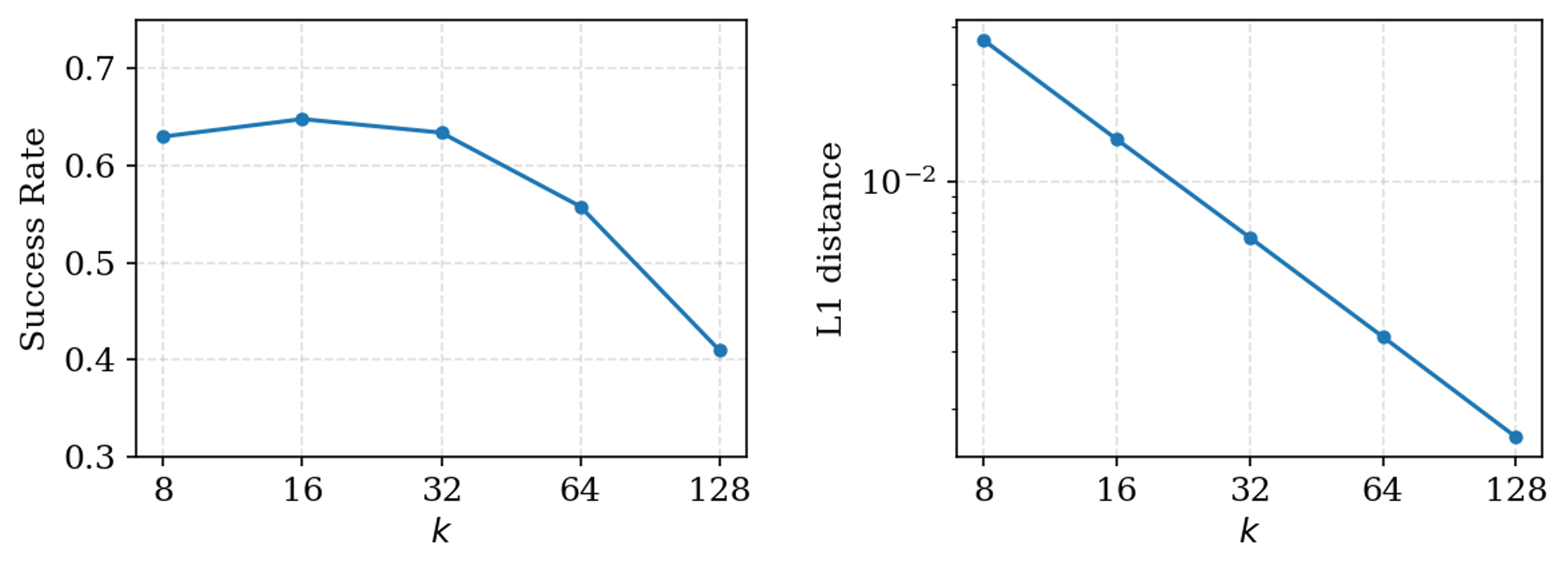}
        \caption{
  \textbf{Left:} Success rate (SR) in the LIBERO environment across discretization levels $k$. 
  \textbf{Right:} L1 distance between the original training data and our transformed dataset for each $k$.}
    \label{fig:libero_kmeans}
\end{figure*}
}

Figure~\ref{fig:libero_kmeans} (left) shows the task success rate for different values of $k$. Interestingly, performance degrades as $k$ increases, despite the more accurate representation of the original continuous action space. This trend is counterintuitive and can be explained by the nature of CLIP-RT’s model design. CLIP-RT formulates action prediction as a classification problem over a fixed set of discrete language-based motion candidates. Increasing $k$ expands the action vocabulary and introduces greater similarity and overlap between candidate motions, making the contrastive learning task more difficult. As a result, the model struggles to confidently select among many fine-grained options, leading to a drop in task performance.

This observation is further supported by Figure~\ref{fig:libero_kmeans} (right), which visualizes the L1 distance between the original continuous actions and their discretized counterparts. As expected, finer quantization (larger $k$) results in lower L1 error. However, lower error does not necessarily lead to better learning under CLIP-RT’s current classification-based framework.

From this study, we draw two key conclusions. First, CLIP-RT’s classification-based approach, while effective for learning from language-supervised data, suffers from information loss when learning from continuous trajectories, achieving only 64.8\% success rate compared to OpenVLA’s 84.7\% on LIBERO (A similar trend is observed on LIBERO-Object, where CLIP-RT reaches 74.8\% while OpenVLA achieves 88.4\%). This represents a limitation of our current framework. Second, as shown in Section~\ref{ssec:libero_main}, it is possible to extend CLIP-RT to handle continuous data more effectively by attaching an action head and training it with regression objectives—this is the basis of CLIP-RT+.

Nevertheless, we believe that language should serve as a natural and accessible interface for robot learning, particularly in scenarios where demonstrations are provided by non-experts. Our approach provides a scalable and intuitive framework for learning language-conditioned robotic policies in such settings. Looking ahead, future work may explore hybrid learning paradigms that combine natural language with complementary modalities—such as spatial cues or proprioceptive feedback—to enable more flexible and robust skill acquisition.

}

\subsection{Runtime Evaluation}
\label{ssec:runtime}
In this section, we provide a comparison of the parameter sizes and runtimes of CLIP-RT and OpenVLA. 
CLIP-RT has a parameter size of 1 billion (1B), while OpenVLA has a significantly larger parameter size of 7 billion (7B).
To evaluate the performance at runtime of each model, we measured the average time taken to process one sample (image and instruction) across 80 samples. 
Each sample begins from the point of receiving the image and instruction until the model returns an action. 
Note that this runtime evaluation excludes the time required for real-world robot actions and server communication, as it focuses solely on the model's processing time.
CLIP-RT achieves an average processing rate of 16Hz when running on a single H100 GPU, demonstrating its efficiency and responsiveness in generating actions. 
In contrast, OpenVLA runs at a slower rate of 2Hz under the same conditions, reflecting its larger parameter size and increased computational demand. 
This runtime performance highlights the efficiency advantage of CLIP-RT, making it well-suited for real-time robotic control and embodied AI applications.

%% file: sections/appx/08_task_details.tex
\section{Task Details}
\label{sec:appendix_dataset}

\begin{figure*}[h]
    \centering
    \includegraphics[width=0.6\textwidth]{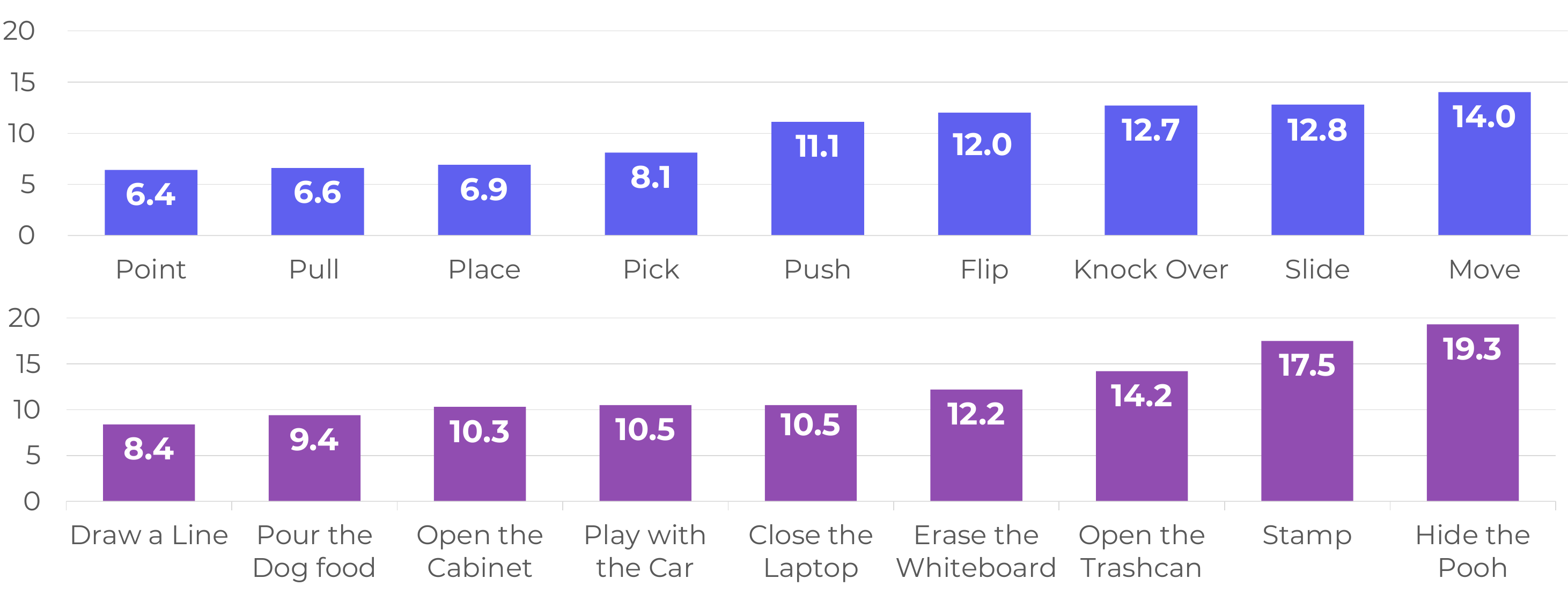}
    \caption{\textbf{Average trajectory length for each task.} The number on the top of each bars represent the average trajectory length of each task, \textit{i.e.}, average number of actions taken. Common tasks are shown at the top, while Novel tasks are shown at the bottom.}
    \label{fig:task_trajectory}
\end{figure*}

In our experiment, we aim to comprehensively evaluate the capabilities of the CLIP-RT model by testing it across both \textit{Common} and \textit{Novel} tasks. 
\textbf{Common tasks} consist of 9 tasks that were part of the pretraining dataset (\textit{i.e.}, Open-X Embodiment dataset~\cite{padalkar2023open}). 
These tasks enable us to evaluate the performance of CLIP-RT, where it benefits from prior knowledge, reflecting its ability to handle familiar tasks.
On the other hand, \textbf{Novel tasks} include 9 previously unseen tasks, which were not part of the pretraining dataset. 
The purpose of novel tasks is to evaluate whether CLIP-RT can learn new skills effectively using our proposed method.
To assess the complexity of the task, we analyze the average number of transitions of each task collected through language-based teleoperation. 
This information provides insight into the varying levels of difficulty, as shown in Figure~\ref{fig:task_trajectory}. 
In addition, we provide detailed descriptions and visual examples of all 19 tasks (Figure~\ref{fig:task_task0} $\sim$ Figure~\ref{fig:task_utask10}).

\subsection{Common Tasks}

\begin{enumerate}[leftmargin=*, label=\textbf{\arabic*}.]
    \item \textbf{Point}: The robot is expected to locate \texttt{<obj>} (\textit{e.g.}, cups with different colors, dice) and move its gripper closer to it.
    \begin{figure}[h]
        \centering
        \includegraphics[width=0.95\columnwidth]{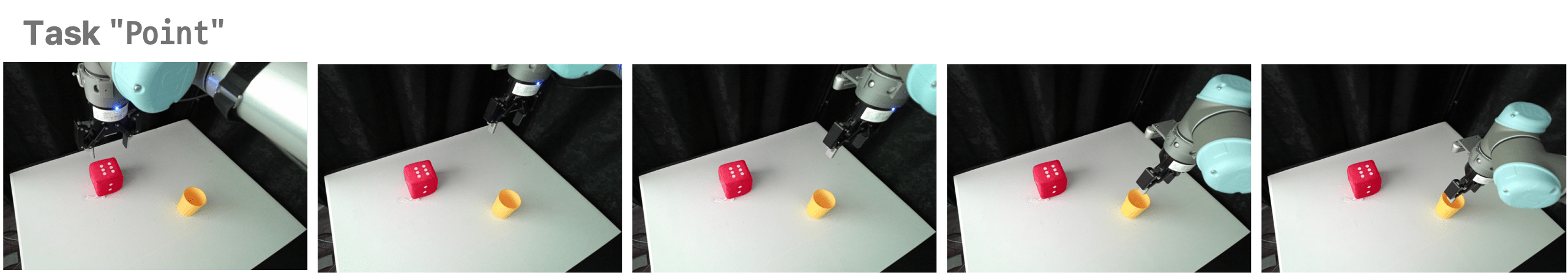}
        \caption{``Point at the yellow cup''}
        \label{fig:task_task0}
    \end{figure} 
    \clearpage

    \item \textbf{Pull}: The robot is required to locate the tissue box, adjust the rotation of its gripper, and grasp and pull the tissue upward. 
    \begin{figure}[h]
        \centering
        \includegraphics[width=0.95\columnwidth]{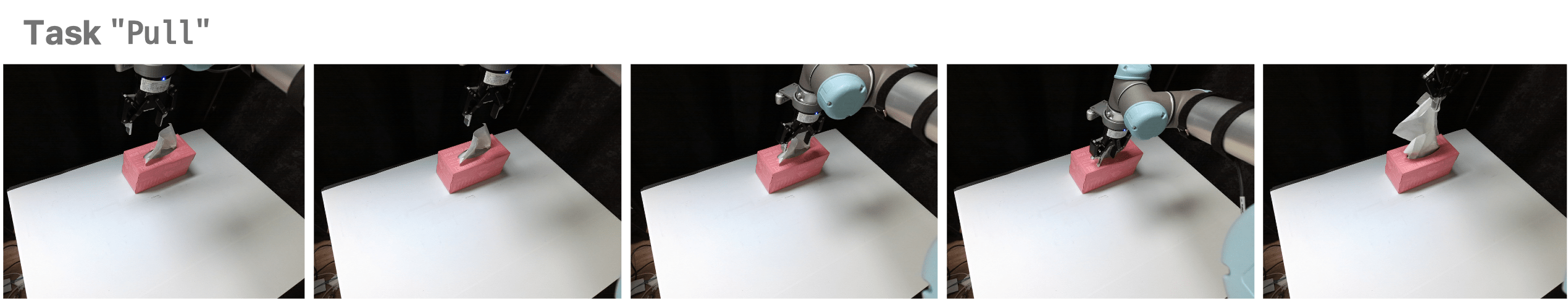}
        \caption{``Pull out the tissue''}
        \label{fig:task_task8}
    \end{figure}

    \item \textbf{Place}: The robot is required to place an object at a designated location (\textit{e.g.}, inside a cabinet, in large colored boxes, or on a shape such as star or circle). The task starts with the robot already grasping the object.
    \begin{figure}[h]
        \centering
        \includegraphics[width=0.95\columnwidth]{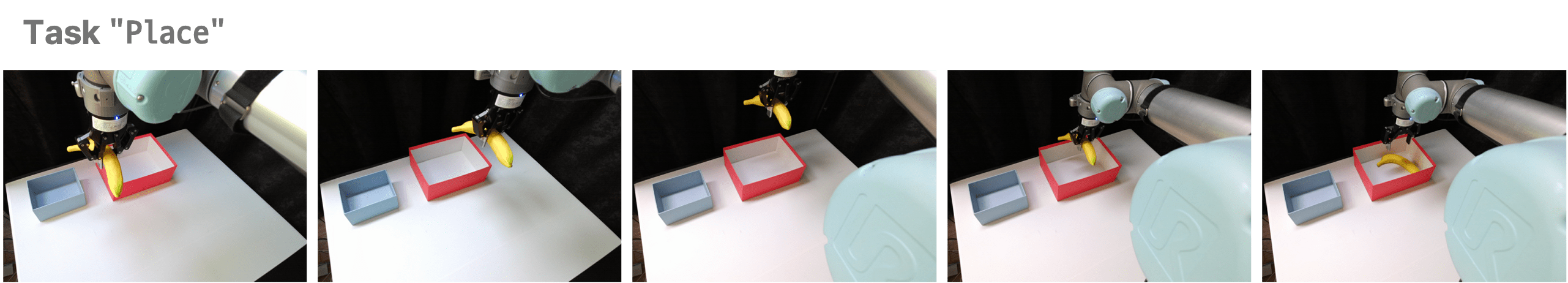}
        \caption{``Place the banana in the red box''}
        \label{fig:task_task2}
    \end{figure}
    
    \item \textbf{Pick}: The robot is expected to find and grasp \texttt{<obj>}. The target objects include cup, dice, and stamp, as well as banana. 
    \begin{figure}[h]
        \centering
        \includegraphics[width=0.95\columnwidth]{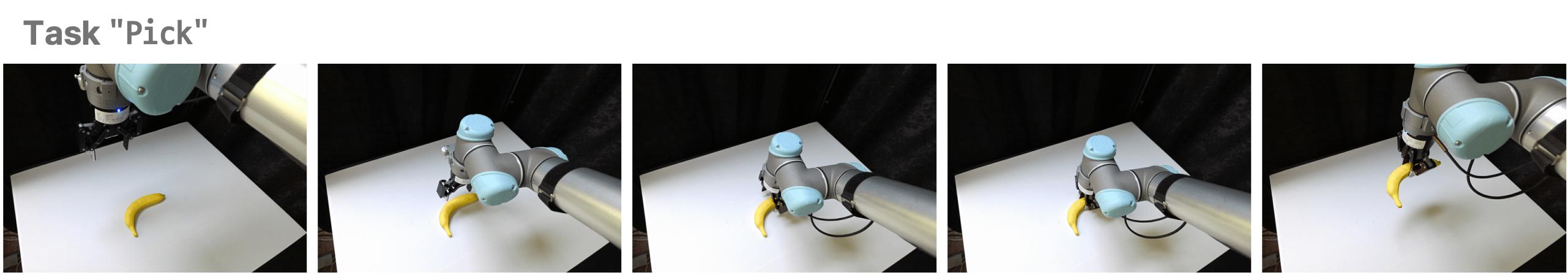}
        \caption{``Pick the banana''}
        \label{fig:task_task1}
    \end{figure}
    
    \item \textbf{Push}: The robot is required to position itself in front of \texttt{<obj1>} (\textit{e.g.}, a dice, box, cup, etc.) and push the object using its arm to move it toward another \texttt{<obj2>}. 
    \begin{figure}[h]
        \centering
        \includegraphics[width=0.95\columnwidth]{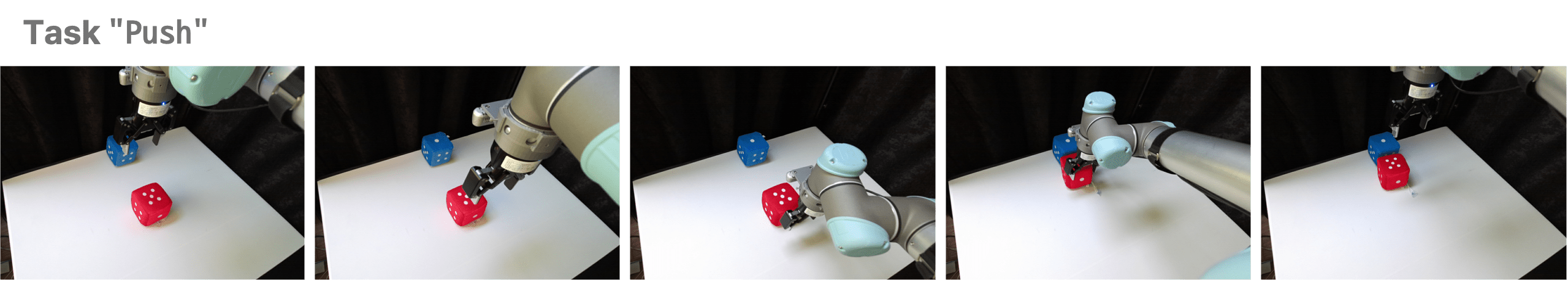}
        \caption{``Push the red dice to the blue dice''}
        \label{fig:task_task3}
    \end{figure}
    
\newpage    
    \item \textbf{Flip}: The robot is required to locate and grasp \texttt{<obj>} (\textit{e.g.}, a cup or a plate), lift it, flip it over by rotating the gripper, and lay it down. 
    \begin{figure}[h]
        \centering
        \includegraphics[width=0.95\columnwidth]{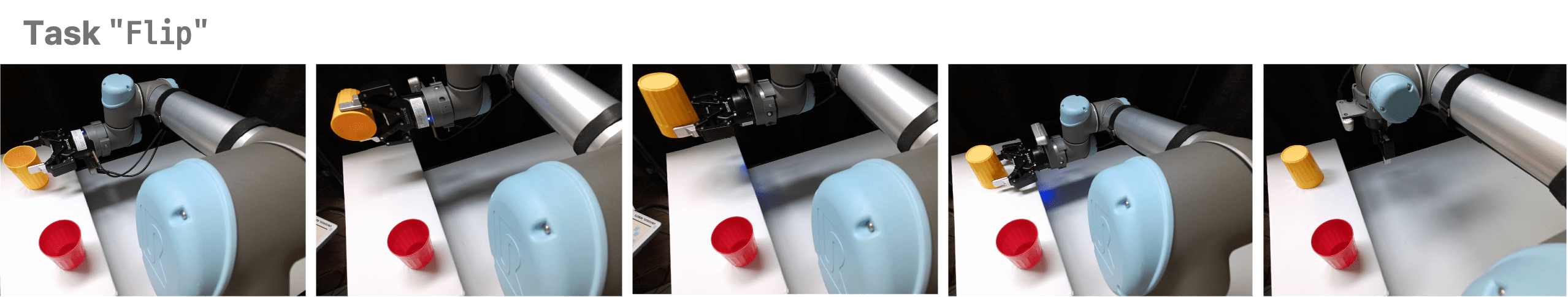}
        \caption{``Flip the yellow cup''}
        \label{fig:task_task7}
    \end{figure}
    
    \item \textbf{Knock Over}: The robot is required to locate and grasp \texttt{<obj>} (\textit{e.g.}, dice, blocks, cups), tilt the object and open the gripper to knock the object over. 
    \begin{figure}[h]
        \centering
        \includegraphics[width=0.95\columnwidth]{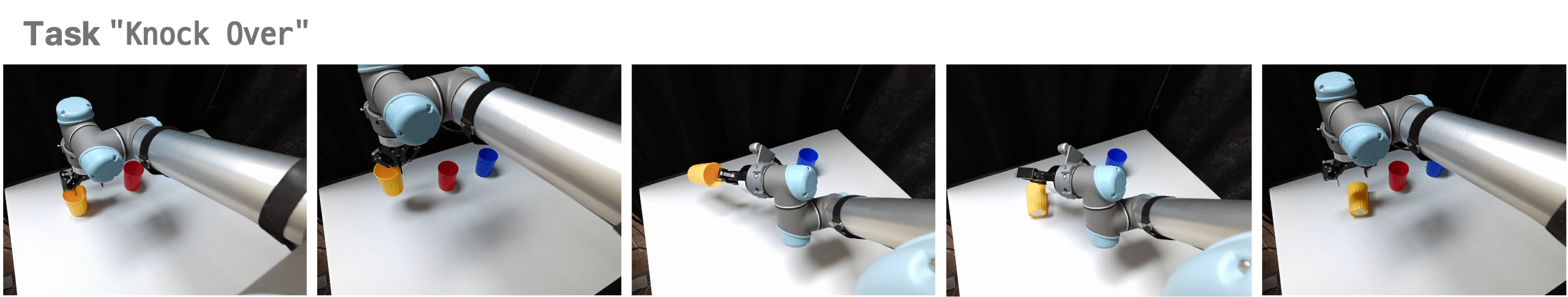}
        \caption{``Knock Over the yellow cup''}
        \label{fig:task_task5}
    \end{figure}
    
    \item \textbf{Slide}: The robot is expected to grasp \texttt{<obj1>} (\textit{e.g.}, dice, toy car) and slide it towards \texttt{<obj2>} (\textit{e.g.}, Pooh, Piglet, or a board eraser). 
    
    \begin{figure}[h]
        \centering
        \includegraphics[width=0.95\columnwidth]{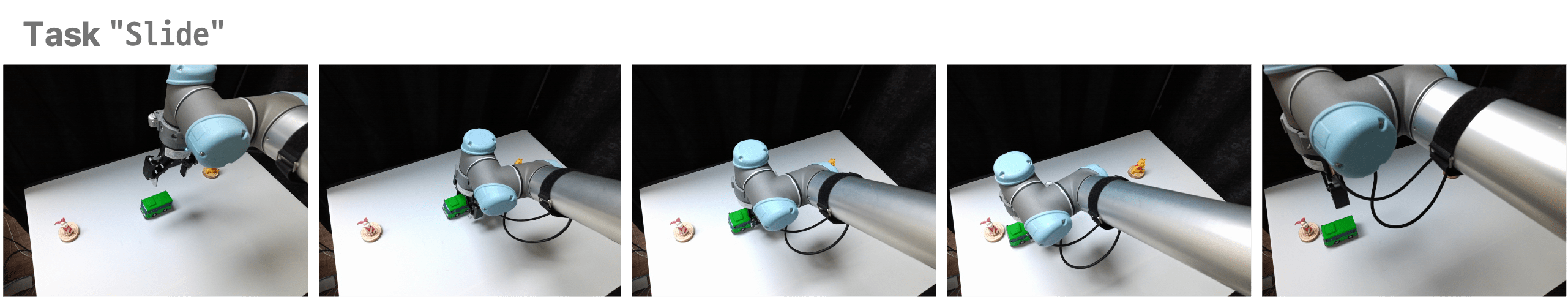}
        \caption{``Slide the red car to the Piglet''}
        \label{fig:task_task6}
    \end{figure}

    \item \textbf{Move}: The robot is required to locate and grasp \texttt{<obj1>} (\textit{e.g.}, banana, cups, plate, etc.), lift it and place it in the designated location (\textit{ e.g.}, near the \texttt{<obj2>}, on the \texttt{<obj2>}). 
    
    \begin{figure}[h]
        \centering
        \includegraphics[width=0.95\columnwidth]{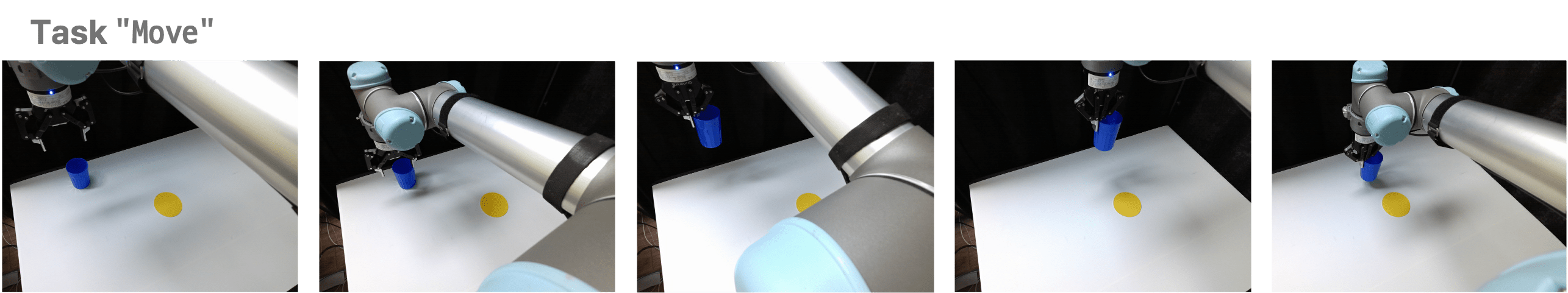}
        \caption{``Move the blue cup near the yellow circle''}
        \label{fig:task_task4}
    \end{figure}
    
\end{enumerate}
\newpage
\subsection{Novel Tasks}

\begin{enumerate}[leftmargin=*, label=\textbf{\arabic*}.]
    \item \textbf{Draw a Line}: The task starts with a board marker held in the robot's gripper. The robot should draw a line that meets the specified condition, such as drawing the line vertically, horizontally, from \texttt{<obj1>} to \texttt{<obj2>}, or between \texttt{<obj1>} and \texttt{<obj2>}. 
    \begin{figure}[h] 
        \centering
        \includegraphics[width=0.95\columnwidth]{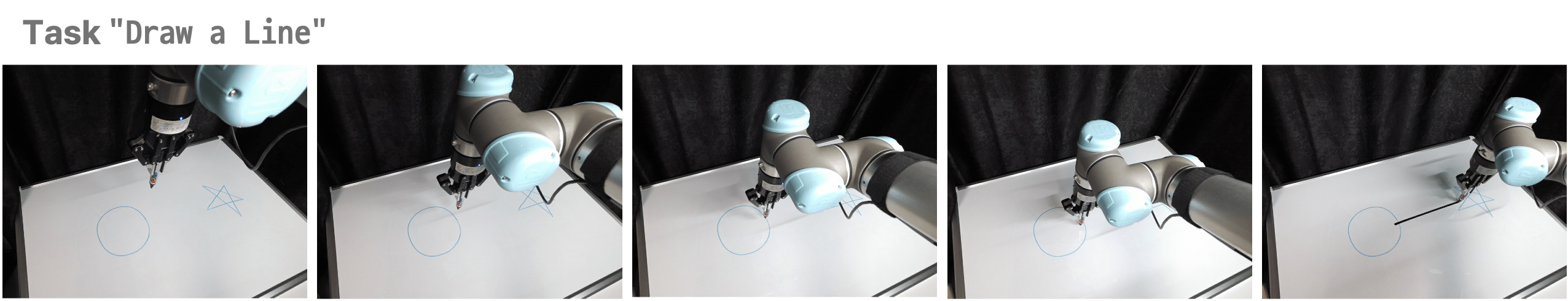}
        \caption{``Draw a Line from the circle to the star''}
        \label{fig:task_utask5}
    \end{figure}

        \item \textbf{Pour the Dog Food}: The robot starts by holding a blue shovel filled with dog food. The robot has to locate the silver bowl and tilt its arm to pour the dog food into the bowl. 
    \begin{figure}[h] 
        \centering
        \includegraphics[width=0.95\columnwidth]{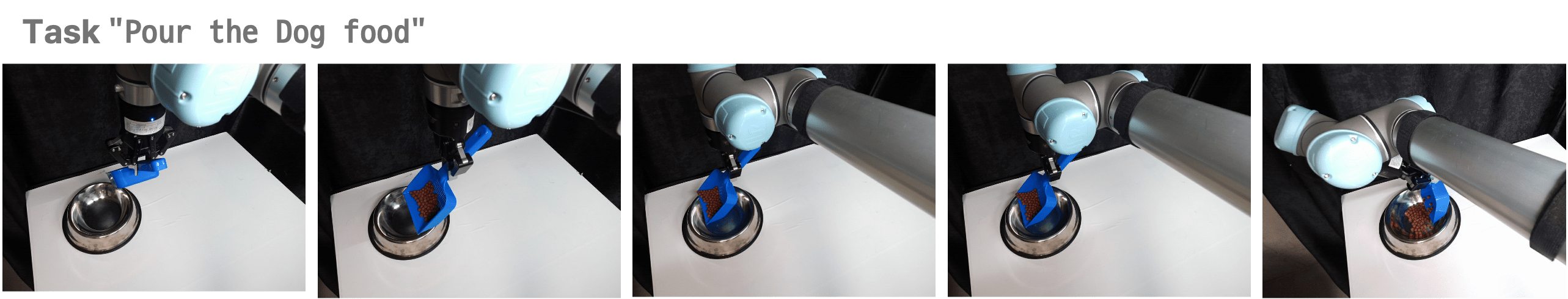}
        \caption{``Pour the Dog food in the bowl''}
        \label{fig:task_utask6}
    \end{figure}
    
    \item \textbf{Open the Cabinet}: The robot is required to position itself in front of the cabinet located in various places. It has to roll or tilt its arm and rotate the gripper to pick up the cabinet handle. 
    \begin{figure}[h] 
        \centering
        \includegraphics[width=0.95\columnwidth]{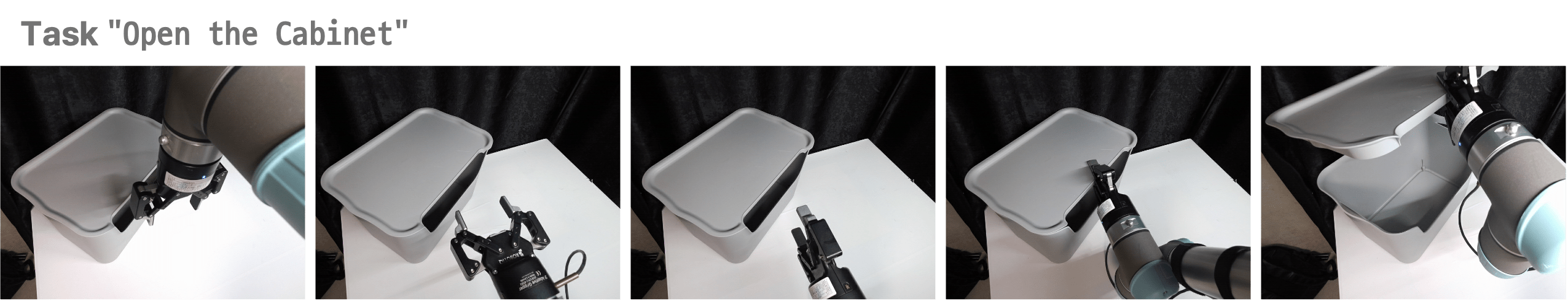}
        \caption{``Open the Cabinet''}
        \label{fig:task_utask1}
    \end{figure}

    \item \textbf{Play with the Car}: The toy car used in the experiment is pull-back car, which moves forward on its own when pulled backward and released. The robot is required to grasp the toy car, move it backward and open the gripper. 
    \begin{figure}[h] 
        \centering
        \includegraphics[width=0.95\columnwidth]{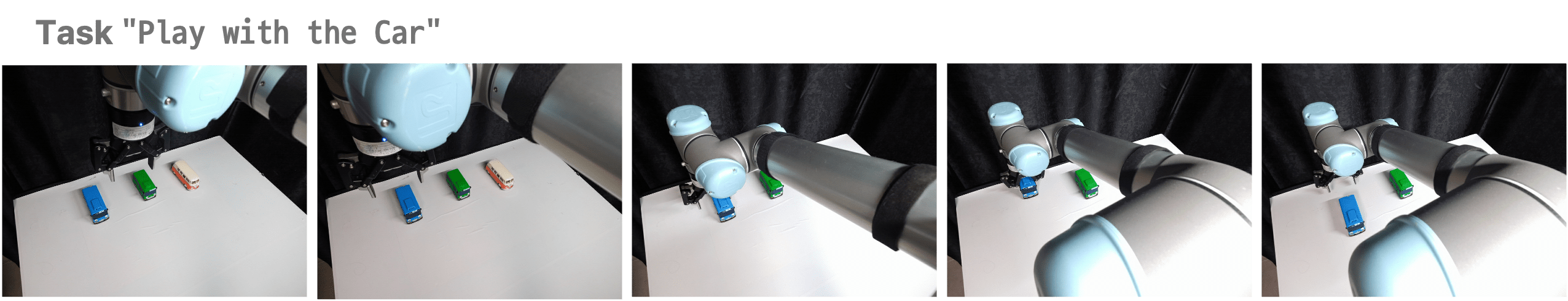}
        \caption{``Play with the blue car''}
        \label{fig:task_utask4}
    \end{figure}

    \clearpage
    \item \textbf{Close the Laptop}: The robot is expected to locate the laptop, position itself behind it, and move forward, right, left, or downward until the laptop is closed. 
    \begin{figure}[h] 
        \centering
        \includegraphics[width=0.95\columnwidth]{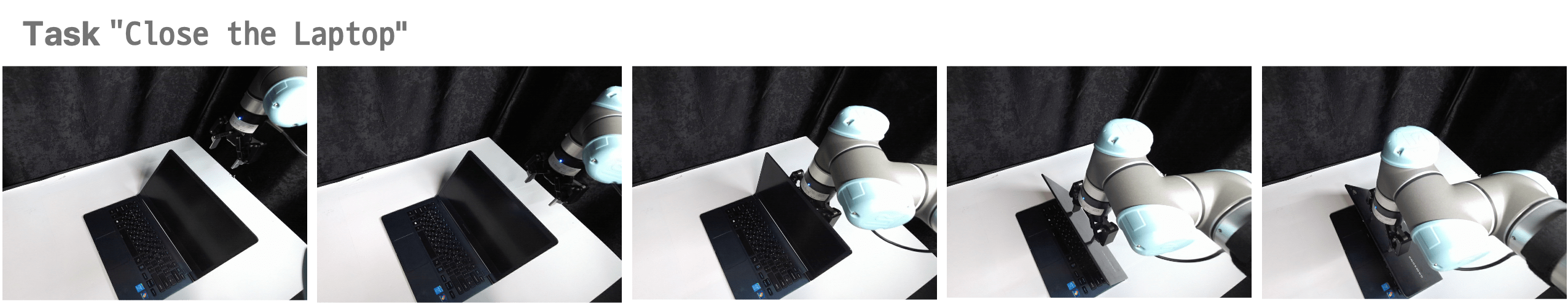}
        \caption{``Close the Laptop''}
        \label{fig:task_utask11}
    \end{figure}

    \item \textbf{Erase the Whiteboard}: The robot is required to locate the doodles on the whiteboard, find and grasp the whiteboard eraser, and slide the eraser to erase the doodles. The doodles may vary in shapes, such as springs or winding lines. 
    
    \begin{figure}[h] 
        \centering
        \includegraphics[width=0.95\columnwidth]{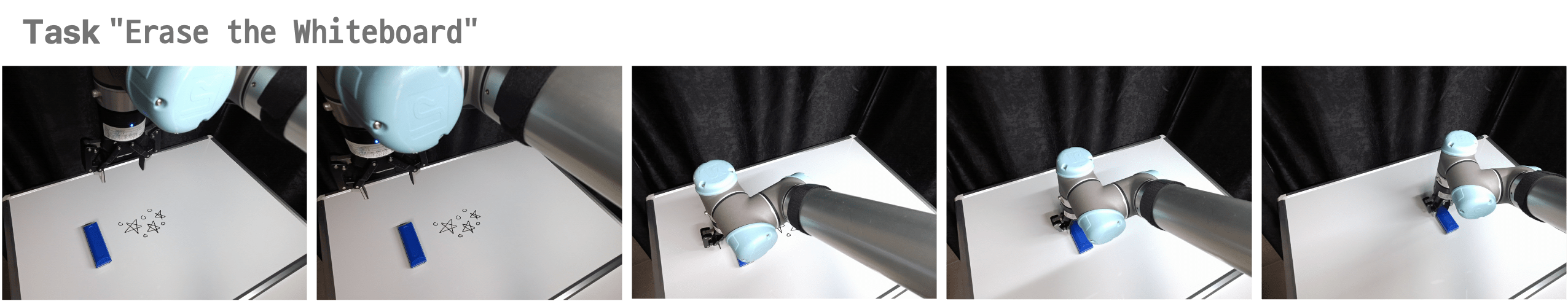}
        \caption{``Erase the Whiteboard''}
        \label{fig:task_utask3}
    \end{figure}
    
    \item \textbf{Open the Trashcan}: The robot is required to approach the press-lid trashcan, precisely press the lid downward and lift its arm to open it. 
    
    \begin{figure}[h] 
        \centering
        \includegraphics[width=0.95\columnwidth]{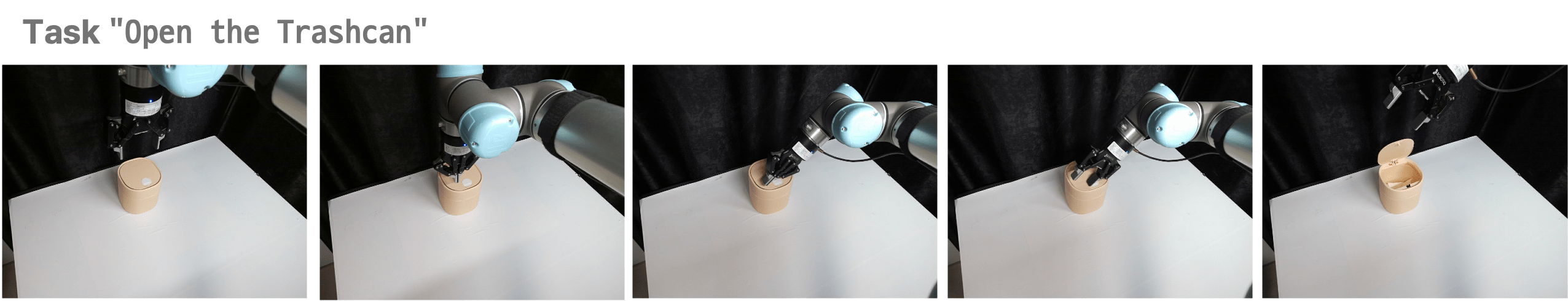}
        \caption{``Open the Trashcan''}
        \label{fig:task_utask2}
    \end{figure}

    \item \textbf{Stamp}: The robot is tasked with grasping the stamp and precisely applying it at the specified position (\textit{e.g.}, next to \texttt{<obj>}, on \texttt{<obj>}, between \texttt{<obj1>} and \texttt{<obj2>}). 

    \begin{figure}[h] 
        \centering
        \includegraphics[width=0.95\columnwidth]{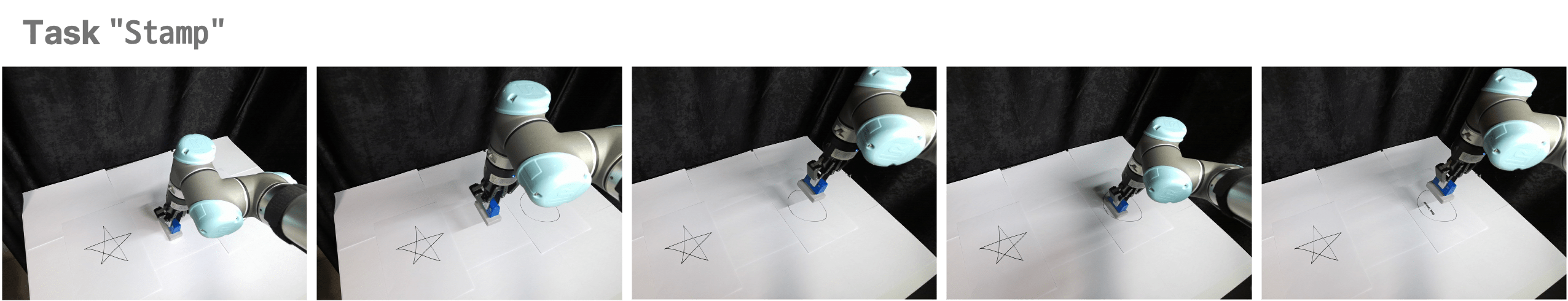}
        \caption{``Stamp on the circle''}
        \label{fig:task_utask7}
    \end{figure}

    \item \textbf{Hide}: The robot must grasp the cup, flip it upside down (if necessary) and place it over the \texttt{<obj>}, ensuring the cup is properly placed to hide the object. The cup may start upside-down from the beginning, and the object to hide could be a toy like Piglet, Pooh, or a small block. 
    \begin{figure}[h] 
        \centering
        \includegraphics[width=0.95\columnwidth]{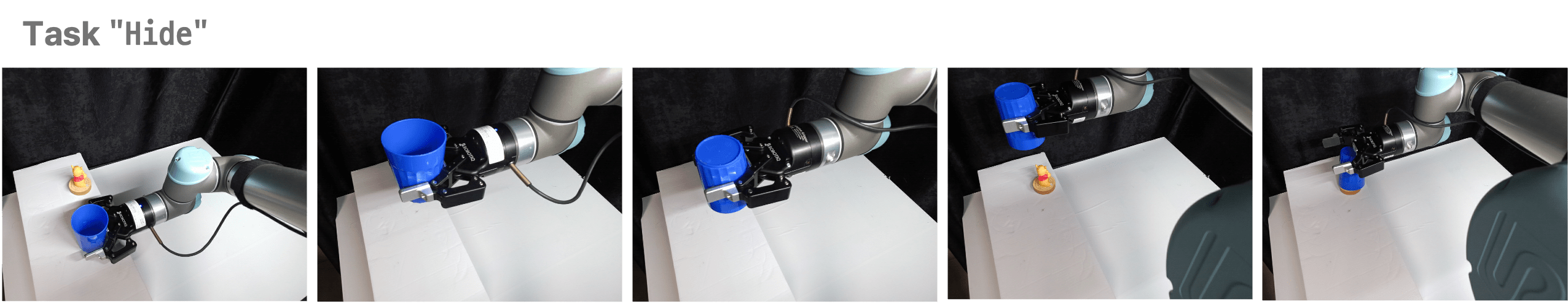}
        \caption{``Hide the Pooh with the blue cup''}
        \label{fig:task_utask10}
    \end{figure}
    
\end{enumerate}


%% file: sections/appx/11_embedding_vis.tex
\section{Embeddings Visualization}
\label{sec:embedding_vis}

Figure~\ref{fig:tsne} shows the t-SNE projection~\cite{van2008visualizing} of 58 motion primitives (see Appendix~\hyperref[tab:lookup]{B}) for CLIP-RT (left) and CLIP-RT-Action (right). We project the vector embeddings of each motion primitive into the 2D space and categorize motion primitives into 16 groups based on the type of displacement. For example, the orange-colored points denote motions about moving the robot arm to the right with different amounts of movement (1cm, 5cm, 10cm, and 20cm). In Figure~\ref{fig:tsne}, CLIP-RT tends to embed the same groups of motions closer, while CLIP-RT-Action does not show clear structures in its embeddings. It implies that natural language supervision enables CLIP-RT to leverage inherent \emph{language priors} in the pretrained vision-language model (\textit{i.e.,} CLIP), facilitating the learning of more structured and semantically meaningful action representations. We conjecture that these language priors improve the generalization capabilities of CLIP-RT.

\begin{figure*}[!h]
    \centering
    \includegraphics[width=\textwidth]{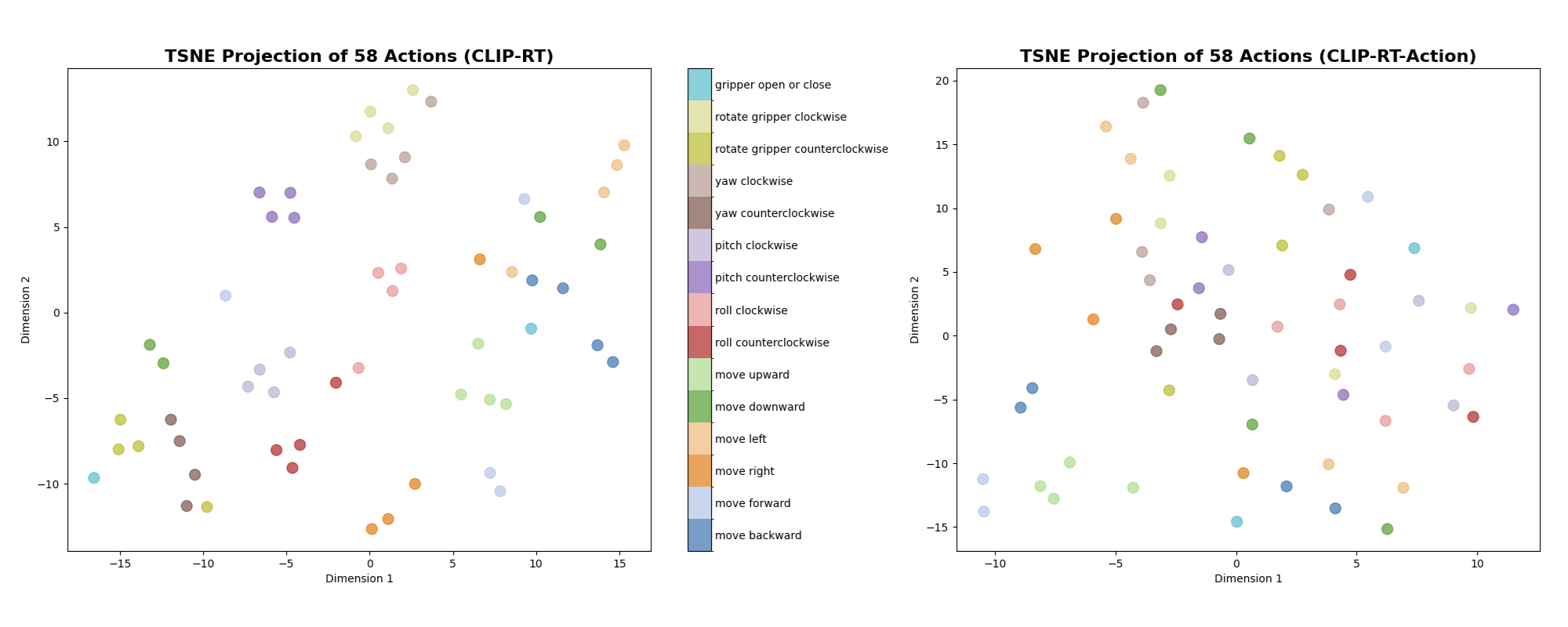}
    \caption{
        \textbf{t-SNE visualization of 58 motion primitive embeddings from CLIP-RT (left) and CLIP-RT-Action (right)}.}
    \label{fig:tsne}
\end{figure*}

%% file: sections/appx/09_ensemble_additional.tex
\section{CLIP-RT $+$ GPT: Details and Examples}

\subsection{CLIP-RT + GPT: Details of Prompt Design and Score Integration}
\label{ssec:prompt_gpt}

To guide GPT in selecting and scoring potential actions, we provide it with both the current image and a language prompt. This prompt directs GPT to categorize a limited set of four candidate actions (\texttt{move arm back}, \texttt{move arm forward}, \texttt{move arm to the right} and \texttt{move arm to the left}) into \emph{Appropriate} or \emph{Inappropriate} lists, based on high-level instruction and visual context. Although we initially considered additional actions (e.g., \emph{move arm up}, \emph{down}, \emph{roll}, \emph{pitch}, \emph{yaw}, or \emph{rotate gripper}), GPT struggled to reason consistently about these expanded options. Consequently, we restrict the candidate actions to four directions. In future work, we plan to refine the prompting mechanism further by providing GPT with richer 3D information, such as object and robot coordinates, to better support more sophisticated cooperation.

\begin{tcolorbox}[colback=black!5!white, colframe=black!50!black]

\textbf{TASK DESCRIPTION:}

You are a generalist agent tasked with controlling a physical robot arm equipped with a two-finger gripper, under natural language supervision from humans. Your objective is to compute and return a Python dictionary with two keys: ``Appropriate" and ``Inappropriate." Each key should have a value that is a Python list. These lists should consist of the most appropriate action the robot should take immediately based on its current position and orientation and the most inappropriate actions it should avoid.\vspace{0.3cm}
    
\textbf{ACTION CANDIDATES:}

[
``move arm back",
``move arm forward",
``move arm to the right",
``move arm to the left"
]
\vspace{0.3cm}

\textbf{ENVIRONMENT SETUP:}

The 3D Cartesian coordinate system of the environment is defined as follows:
1. The x-axis is in the depth direction, increasing away from the robot.
2. The y-axis is in the horizontal direction, increasing to the left.
\vspace{0.3cm}

\textbf{RULES:}

\begin{enumerate}[topsep=0pt, partopsep=0pt, itemsep=0pt, parsep=0pt]
\item Only output a single action for the ``Appropriate" key and multiple actions for the ``Inappropriate" key.
\item Align the robot arm horizontally (right and left) first.
\end{enumerate}
\vspace{0.3cm}

\textbf{OUPUT FORMAT:}

Provide a Python dictionary with the keys ``Appropriate" and ``Inappropriate." Appropriate should map to a list of the most appropriate action, and Inappropriate should map to a list of the most inappropriate actions.
\vspace{0.3cm}

\textbf{Instruction}: $\left\{\text{instruction}\right\}$.

Which way should the robot move?
\end{tcolorbox}

Based on prompt design, GPT produces two sets of actions at each time step: \emph{Appropriate} and \emph{Inappropriate}. We incorporate these decisions into the original CLIP-RT action scores as follows. For each action labeled \emph{Appropriate}, we multiply the CLIP-RT score by \((1 + \alpha^t)\); for each \emph{Inappropriate} action, we multiply by \((1 - \alpha^t)\). Here, \(\alpha\) serves as a \emph{guidance factor}, and \(t\) represents the current time step (the number of transitions since the task began).
In this exerpiment, we set \(\alpha = 0.7\). 
By \(t=2\), \emph{Appropriate} actions receive a factor of approximately \(1.5\), significantly influencing which action is ultimately selected. After about ten steps, \(\alpha^t\) becomes negligible, reducing the influence of GPT in later stages. 
The rationale is that GPT’s broader contextual reasoning is particularly helpful in the early stages, while CLIP-RT's precise control is more valuable once the robot is closer to the target object.
Finally, we take the action with the highest adjusted score as the final output at each time step. This strategy allows GPT to provide initial high-level guidance while still using CLIP-RT’s fine-grained capabilities as the task progresses.

\subsection{CLIP-RT + GPT: Qualitative Results}

\label{ssec:appendix_ensemble}

\begin{figure*}[!t]
    \centering
    \includegraphics[width=1\textwidth]{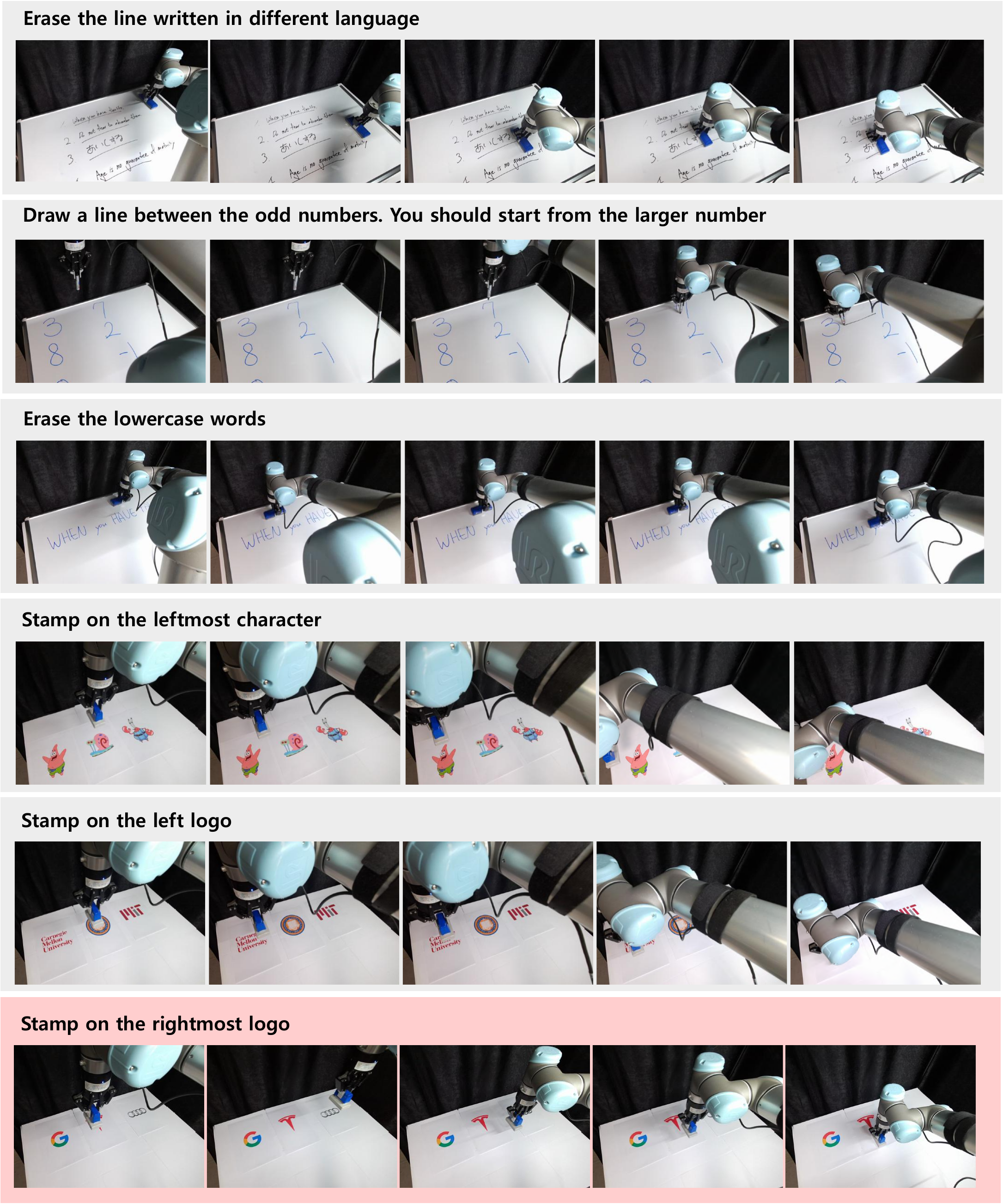}
    \caption{
        \textbf{Qualitative Examples of CLIP-RT + GPT: High-level Reasoning Instructions}    
    }
    \label{fig:ensemble_ex1_hl}
\end{figure*}

\begin{figure*}[!t]
    \centering
    \includegraphics[width=1\textwidth]{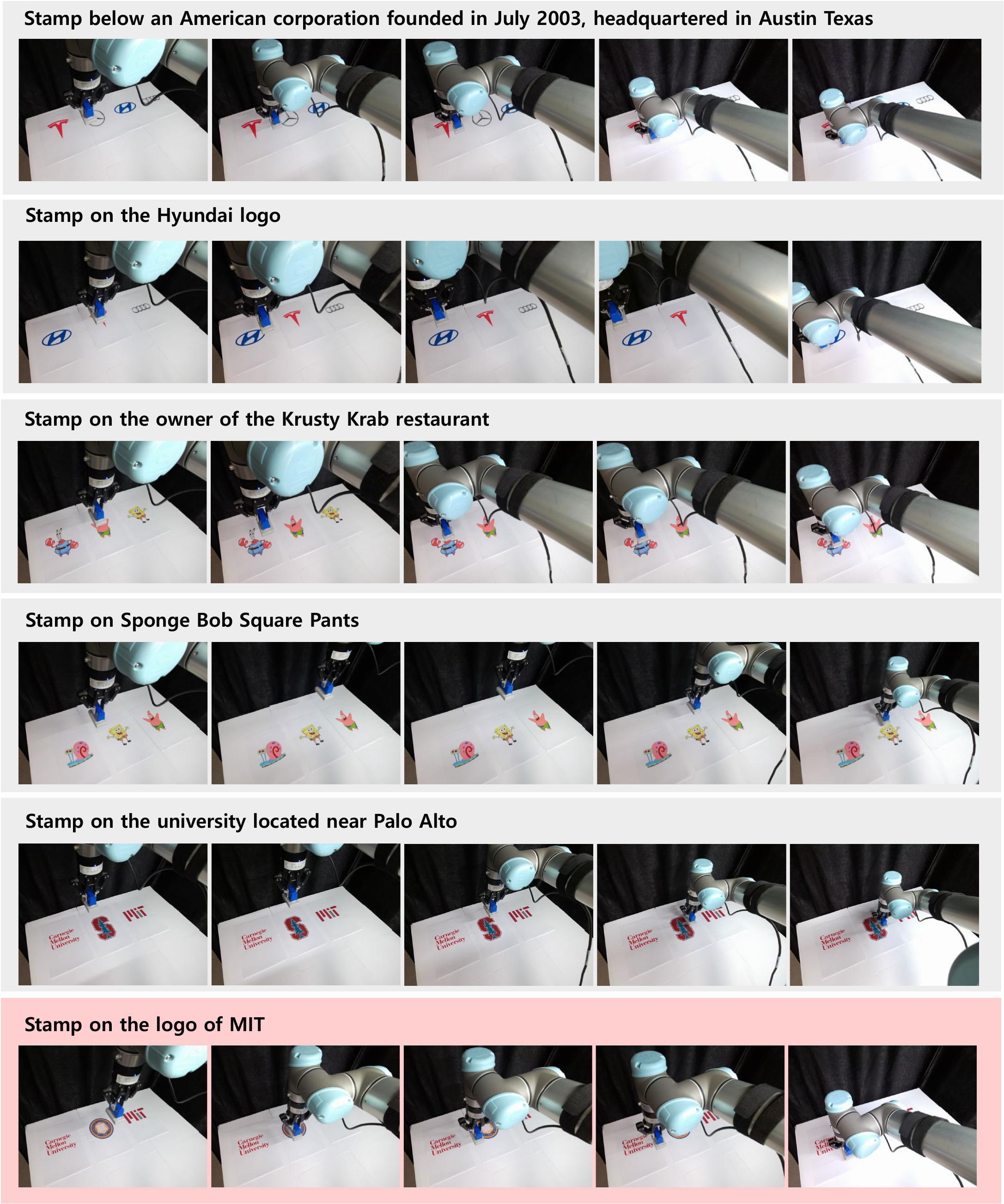}
    \caption{
        \textbf{Qualitative Examples of CLIP-RT + GPT: Knowledge-Intensive Instructions}    
    }
    \label{fig:ensemble_ex2_ki}
\end{figure*}

GPT-guided approach broadens the range of instructions CLIP-RT can handle, enabling it to tackle knowledge-intensive or high-level reasoning scenarios it has not seen during training.
Here we provide qualitative examples that demonstrate instructions that require complex reasoning (Figure~\ref{fig:ensemble_ex1_hl}) and intensive knowledge (Figure~\ref{fig:ensemble_ex2_ki}).
Failure cases are highlighted in red boxes. Because our setup relies on a single camera view and lacks historical context, the target object may become obscured by the robot arm, leading to mispredictions. For instance, in Figure~\ref{fig:ensemble_ex1_hl} (“Stamp on the rightmost logo”), the Audi logo is hidden by the arm, leading the model to mistakenly identify Tesla as the rightmost logo. In future work, incorporating multiple viewpoints or tracking historical states could mitigate this limitation.
However, qualitative examples demonstrate that the collaboration between GPT and CLIP-RT enables the system to effectively execute a diverse range of instructions, showcasing its potential to follow challenging instructions.

